\documentclass{article}

\usepackage[preprint]{neurips_2022}

\usepackage{float}
\usepackage[utf8]{inputenc} 
\usepackage[T1]{fontenc}    
\usepackage{hyperref}       
\usepackage{url}            
\usepackage{booktabs}       
\usepackage{amsfonts}       
\usepackage{nicefrac}       
\usepackage{microtype}      
\usepackage{graphicx}
\usepackage{subfigure}
\usepackage[dvipsnames]{xcolor}
\usepackage{pgfplots}
\usepgfplotslibrary{groupplots}
\usepackage{color}
\pgfplotsset{compat=1.5}

\usepackage{subfigure}
\usepackage{multirow}
\usepackage{graphicx}
\usepackage{ulem}
\usepackage{amsmath}
\usepackage{amssymb,amsthm}
\usepackage{enumitem}
\setlist{nolistsep}

\usepackage{natbib}
\setcitestyle{numbers,square,citesep={,}}

\definecolor{ForestGreen}{RGB}{34,139,34}

\title{Open-Vocabulary 3D Detection via Image-level Class and Debiased Cross-modal Contrastive Learning}

\author{
  Yuheng Lu$^{1*}$, Chenfeng Xu$^2$\thanks{Equal contribution. Corresponding author: Shanghang Zhang, E-mail: \tt\small shanghang@pku.edu.cn} ,    Xiaobao Wei$^1$,    Xiaodong Xie$^1$\\ \textbf{Masayoshi Tomizuka$^2$,    Kurt Keutzer$^2$,    Shanghang Zhang$^1$}\\
  $^1$Peking University,    $^2$UC Berkeley \\
  $^1$$\left\{ yuhenglu, xiaobao, donxie, shanghang \right\}$@pku.edu.cn\\
  $^2$$\left\{ xuchenfeng, tomizuka, keutzer \right\}$@berkeley.edu.
}

\begin{document}

\maketitle

\begin{abstract}
Current point-cloud detection methods have difficulty detecting the open-vocabulary objects in the real world, due to their limited generalization capability. Moreover, it is extremely laborious and expensive to collect and fully annotate a point-cloud detection dataset with numerous classes of objects, leading to the limited classes of existing point-cloud datasets and hindering the model to learn general representations to achieve open-vocabulary point-cloud detection. As far as we know, we are the first to study the problem of open-vocabulary 3D point-cloud detection. Instead of seeking a point-cloud dataset with full labels, we resort to ImageNet1K to broaden the vocabulary of the point-cloud detector. We propose OV-3DETIC, an \textbf{O}pen-\textbf{V}ocabulary \textbf{3}D \textbf{DET}ector using \textbf{I}mage-level \textbf{C}lass supervision. Specifically, we take advantage of two modalities, the image modality for recognition and the point-cloud modality for localization, to generate pseudo labels for unseen classes. Then we propose a novel debiased cross-modal contrastive learning method to transfer the knowledge from image modality to point-cloud modality during training. Without hurting the latency during inference, OV-3DETIC makes the point-cloud detector capable of achieving open-vocabulary detection. Extensive experiments demonstrate that the proposed OV-3DETIC achieves at least 10.77 \% mAP improvement (absolute value) and 9.56 \% mAP improvement (absolute value) by a wide range of baselines on the SUN-RGBD dataset and ScanNet dataset, respectively. Besides, we conduct sufficient experiments to shed light on why the proposed OV-3DETIC works.

\end{abstract}

\section{Introduction}

3D point-cloud detection is defined as finding objects (localization) in point-cloud and recognizing them (classification). Recently, deep learning based 3D detectors have achieved significant progress. However, most methods are developed on point-cloud detection datasets with limited classes (vocabularies), while the real world owns a cornucopia of classes. It is common for 3D detectors to encounter objects that had never occurred during training, resulting in failure to generalize to real-life scenarios. Therefore, it is extremely important to design an open-vocabulary point-cloud detector which generalizes to unseen classes.
The key ingredient of open-vocabulary detection is that the model learns sufficient knowledge thus is able to output general representations. To achieve this, in the image field, typical open-vocabulary classification and detection either require to introduce large-scale image-text pairs or image datasets with sufficient labels. For example, CLIP \cite{radford2021learning} introduced 400 million image-text pairs for pre-training to help visual models learn general representation. Detic \cite{zhou2022detecting} leverages ImageNet21K to extend the knowledge of image detectors. OVR-CNN \cite{zareian2021open} uses
the language pretrained embedding layer to broaden the vocabulary of the 2D detector.

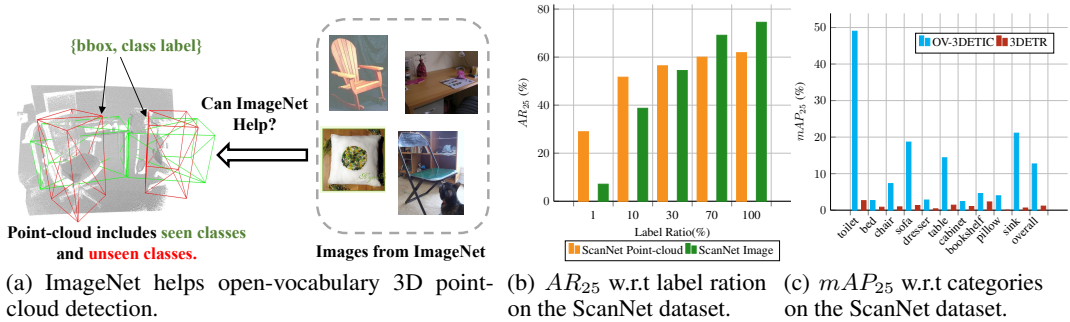
\begin{figure*}[!t]
    \centering
	\subfigure[ImageNet helps open-vocabulary 3D point-cloud detection.]{\label{fig:imagenethelppc}\includegraphics[width=0.46\textwidth]{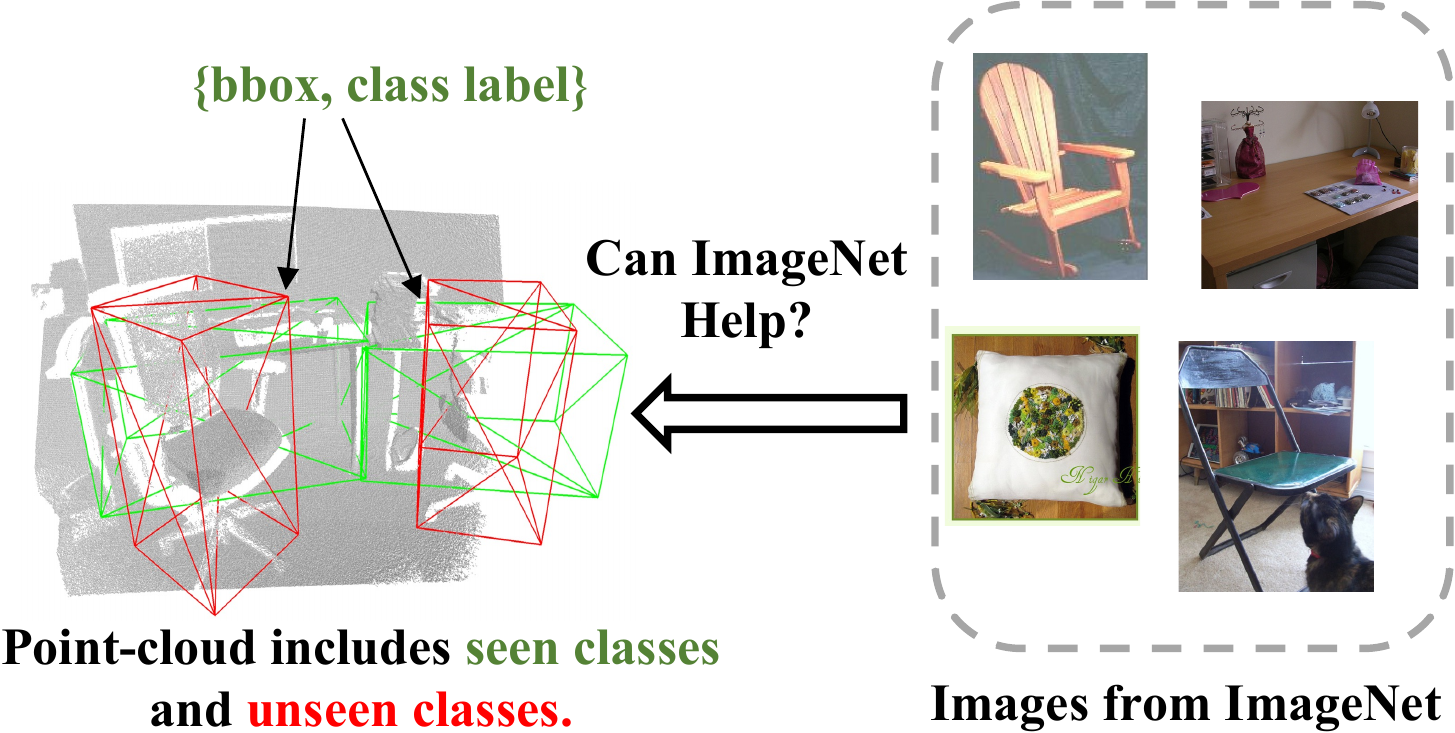}}%
	\hspace{1.5mm}
    \subfigure[$AR_{25}$ w.r.t label ration on the ScanNet dataset.]{
        \label{fig:localization_veryfy}
        \begin{minipage}[t]{0.24\linewidth}
            \centering
            \begin{tikzpicture}[scale=0.46]
                \centering
            	\begin{axis}[
            	    symbolic x coords={1,10,30,70,100,0},
                    xtick={1,10,30,70,100,0},
            	    x tick label style={rotate=0, anchor=north, align=center},
            	    ylabel=$AR_{25}$ (\%),
            	    xlabel=Label Ratio(\%),
            	    axis lines*=left,
            	    ymajorgrids = true,
            	    ymin = 0,
            	    ybar,
            	    ybar interval=0.5,
            	    legend style={at={(0.5,-0.2)},anchor=north,legend columns=-1},
            	]
                    \addlegendentry{ScanNet Point-cloud}
                    \addplot[draw=BurntOrange, fill=BurntOrange] 
                    coordinates {
                        (1,28.93)
                        (10,51.65)
                        (30,56.38)
                        (70,59.98)
                        (100,61.84)
                        (0,0)
                    };
                    
                    \addlegendentry{ScanNet Image}
                    \addplot[draw=ForestGreen, fill=ForestGreen] 
                    coordinates {
                        (1,7.1)
                        (10,38.7)
                        (30,54.4)
                        (70,69.1)
                        (100,74.5)
                        (0,0)
                    };
                \end{axis}
            \end{tikzpicture}
        \end{minipage}%
    }%
    \hspace{1.5mm}
    \subfigure[$mAP_{25}$ w.r.t categories on the ScanNet dataset.]{
        \label{fig:map_performance}
        \begin{minipage}[t]{0.24\linewidth}
            \centering
            \begin{tikzpicture}[scale=0.46]
                \centering
            	\begin{axis}[
            	    symbolic x coords={toilet, bed, chair, sofa, dresser, table, cabinet,bookshelf,pillow,sink,overall,0},
            	    xtick={toilet, bed, chair, sofa, dresser, table, cabinet,bookshelf,pillow,sink,overall,0},
            	    x tick label style={rotate=45, anchor=east, align=center},
            	    ylabel=$mAP_{25}$ (\%),
            	    axis lines*=left,
            	    ymajorgrids = true,
            	    ymin = 0,
            	    ybar,
            	    ybar interval=0.5,
            	    legend style={at={(0.65,0.9)},anchor=north,legend columns=-1},
            	]
                    \addlegendentry{OV-3DETIC}
                	\addplot[draw=ProcessBlue, fill=ProcessBlue] 
                	coordinates{
                		(toilet,48.99)
                        (bed,2.63)
                        (chair,7.27)
                        (sofa,18.64)
                        (dresser,2.77)
                        (table,14.34)
                        (cabinet,2.35)
                        (bookshelf,4.54)
                        (pillow,3.93)
                        (sink,21.08)
                        (overall,12.65)
                        (0,0)
                	};
                	 
                	\addlegendentry{3DETR}
                	\addplot[draw=BrickRed, fill=BrickRed] 
                	coordinates{
                		(toilet,2.60)
                        (bed,0.81)
                        (chair,0.90)
                        (sofa,1.27)
                        (dresser,0.36)
                        (table,1.37)
                        (cabinet,0.99)
                        (bookshelf,2.25)
                        (pillow,0)
                        (sink,0.59)
                        (overall,1.11)
                        (0,0)
                	};
                \end{axis}
            \end{tikzpicture}
        \end{minipage}%
    }%
	\caption{(a). The left point-cloud includes the tables (green bounding box) and the chairs (red bounding box). The tables are labeled and denoted as seen classes, while the chairs are not annotated and denoted as unseen classes. Right part shows examples from ImageNet1K which has sufficient labels. We aim at utilizing large-scale ImageNet1K to broaden the vocabulary of point-cloud detector. (b). We train the 3DETR \cite{misra2021end} and DETR \cite{carion2020end} on the ScanNet dataset. Under low-data regime, e.g., with randomly sampling 10 \% data, $AR_{25}$ is still at a high-level accuracy, even competitive to using 100 \% data, which demonstrates that localization in point-cloud object detection generalizes well. (c). The performance comparison between the proposed OV-3DETIC and the baseline 3DETR on the ScanNet dataset. OV-3DETIC significantly improves the baseline by a large margin on all the categories. }
	\vspace{-5mm}
\end{figure*}

    

However in the point-cloud field, as far as we know, there is no existing study for the open-vocabulary point-cloud detection. The most notable hindrance is that we can hardly obtain large-scale point-cloud data and labels (or optionally, captions like the aforementioned image field), due to the difficulty of collection and annotation. The scarcity of point-cloud data and labels drastically restricts the point-cloud model from learning sufficient knowledge and obtaining general representations. Therefore, this limitation motivates us to ask - can we transfer the knowledge from images to point-cloud so that the point-cloud model is capable of learning general representations? Our work shows that the answer is positive. Intrinsically, images are dense RGB pixels while point-clouds consist of sparse xyz points. Although the large gap exists, both point-cloud and image are visual representations to the physical world and can express the same visual concepts \cite{xu2021image2point,park2022detmatch}. Human-beings have no problem understanding both modalities. But two issues still exist: what kind of image data can we use and how to use the image data?

It is straightforward that we directly utilize the image detection dataset with fully bounding boxes and class labels, and transfer the knowledge from the 2D detector to the point-cloud detector. However, bounding-box level annotation is still laborious and difficult to scale, while open-vocabulary detection requires rich labels to help the detector learn sufficient knowledge \cite{zhou2022detecting}. Therefore, in this paper, instead of seeking to build large-scale point-cloud dataset or using 2D detection dataset, we open-up another path by resorting to large-scale image dataset with image-level class, ImageNet1K \cite{krizhevsky2012imagenet}, to enable the point-cloud detector capable of learning general representations, thus broadening the vocabulary of the point-cloud detector, as shown in Figure. \ref{fig:imagenethelppc}. As far as we know, we are the first to study the problem of open-vocabulary point-cloud detection. 

Specifically, we propose a \textbf{O}pen-\textbf{V}ocabulary \textbf{3}D \textbf{Det}ector with \textbf{I}mage-level \textbf{C}lass, termed as OV-3DETIC, which aims at making the 3D detector learning sufficient knowledge from image-level supervision thus achieving open-vocabulary point-cloud detection. Remarkably, we observe that the "localization" of point-cloud detector is already able to generalize to unseen categories although the corresponding supervisions are never used during the training, as shown in Figure. \ref{fig:localization_veryfy}. Nevertheless, the "classification" of the point-cloud detector can hardly generalize to unseen classes. This motivates us to take advantage of two modalities during training, the image modality for classification and the point-cloud modality for localization. Particularly, OV-3DETIC is a synergy of two components: 1) Make full use of knowledge learning from ImageNet and generalizability of localization on point-cloud to generate pseudo labels for unseen class. 2) We design a debiased cross-modal contrastive learning with distance-aware temperature to capture the shared low-dimensional space within and across modalities, thus better transferring the sufficient knowledge from image domain to point-cloud domain. It is noteworthy to mention that during training, we introduce the paired images to narrow the gap between the point-cloud data and the images from ImageNet, but we do not need any extra annotations except the Lidar-Camera transformation matrix.

Extensive experiments show that OV-3DETIC outperforms a wide range of state-of-the-arts by at least 10.77\% mAP (absolute) and 9.56 \% mAP (absolute) without hurting the latency of the original 3D detector, on the unseen classes of SUN RGB-D~\cite{song2015sun} and the ScanNet~\cite{dai2017scannet}, respectively. An example on the ScanNet dataset is shown in Figure. \ref{fig:map_performance}. Sufficient ablation studies also shed light on why the OV-3DETIC works. Overall, we summarize the contribution as follows: 
1) We are the first to study the open-vocabulary 3D detection. By discovering the strong generalizability of localization in 3D object detection, we are also the first to open-up another path by resorting to ImageNet1K to help open-vocabulary 3D detection.
2) We propose an open-vocabulary 3D detector with image-level class, termed as OV-3DETIC, which is a synergy of two components: the pseudo-label generation from two modalities, and the debiased cross-modal contrastive learning with distance-aware temperature.
3) Extensive experiments demonstrate the effectiveness of OV-3DETIC. We provide a superior baseline in this field, and also analyse why it works via sufficient experiments.

\section{Related work}
\subsection{Open-Vocabulary Detection on Image.}
Open-vocabulary object detection targets to detect the novel classes that are never provided labels during the training \cite{bansal2018zero,gu2021zero,rahman2020improved,rahman2020zero,zhou2022detecting,zareian2021open,radford2021learning}. The classic open-vocabulary object detection method directly replaces the classifier with language embedding layer \cite{bansal2018zero}. To advance the embedding layer, more popular approaches aim at leveraging image-text pairs to extract the rich semantics from text thus broadening the detector \cite{radford2021learning,gu2021zero,zareian2021open}. 
The most similar work to us is Detic \cite{zhou2022detecting}, which utilizes ImageNet21K to broaden the classifier of the 2D detector. Yet, it is infeasible to directly use the same method to broaden the classifier of the point-cloud detector, due to the large gap between the image and point-cloud. Different from Detic which transfers knowledge from ImageNet with the image-level class to 2D detection within the same modality, we propose to transfer the knowledge from ImageNet to a totally different modality, point-cloud, with customized pseudo-label strategy and debiased cross-modal contrastive learning.

\subsection{Point-Cloud Detection}
Early works on point-cloud object detection discretize and project points onto Bird's Eye View (BEV) or front-view images, and process 2D Lidar feature using standard 2D CNN networks, such as PIXOR~\cite{yang2018pixor}, MV3D~\cite{chen2017multi}, SqueezeSeg~\cite{wu2017squeezeseg,wu2018squeezesegv2,xu2020squeezesegv3}. A more natural way is to directly process each point using PointNet-alike backbones such as PointRCNN~\cite{shi2019pointrcnn} and PointFusion~\cite{xu2017pointfusion}, which is, however, limited by its high computation cost \cite{xu2021you}. Recent popular method is the voxel representation~\cite{zhou2017voxelnet}, which can not only be processed efficiently using 3D sparse convolution \cite{yan2018second,shi2020part,shi2020pv}, but also preserve approximately similar information to raw point cloud with small voxel size. Recently, vision transformer dominates the field of image field~\cite{dosovitskiy2020image,wu2020visual,liu2021swin}, and point-cloud transformer is also gradually developed \cite{misra2021end,zhao2021point}. Our method is based on 3DETR \cite{misra2021end}.

\subsection{Zero-Shot Learning in Point-cloud}
Previous zero-shot (open-vocabulary) learning works in the point-cloud field mainly study classification. Image2Point \cite{xu2021image2point} directly inflates the 2D model pre-trained on large-scale image dataset, and shows a significant improvement for point-cloud classification. PointCLIP \cite{zhang2021pointclip} leverages CLIP pre-trained embedding to broaden vocabulary of the point-cloud classifier. In \cite{cheraghian2019zero,cheraghian2019mitigating,cheraghian2021zero}, PointNet is pre-trained on seen classes and classifies unseen objects by calculating the similarity to the seen class. Recently, although zero-shot semantic segmentation in point-cloud is studied \cite{liu2021segmenting,michele2021generative}, as far as we know, we are the first to explore the open-vocabulary point-cloud detection.

\begin{figure*}[!t]
    \centering
    \includegraphics[width=.80\textwidth]{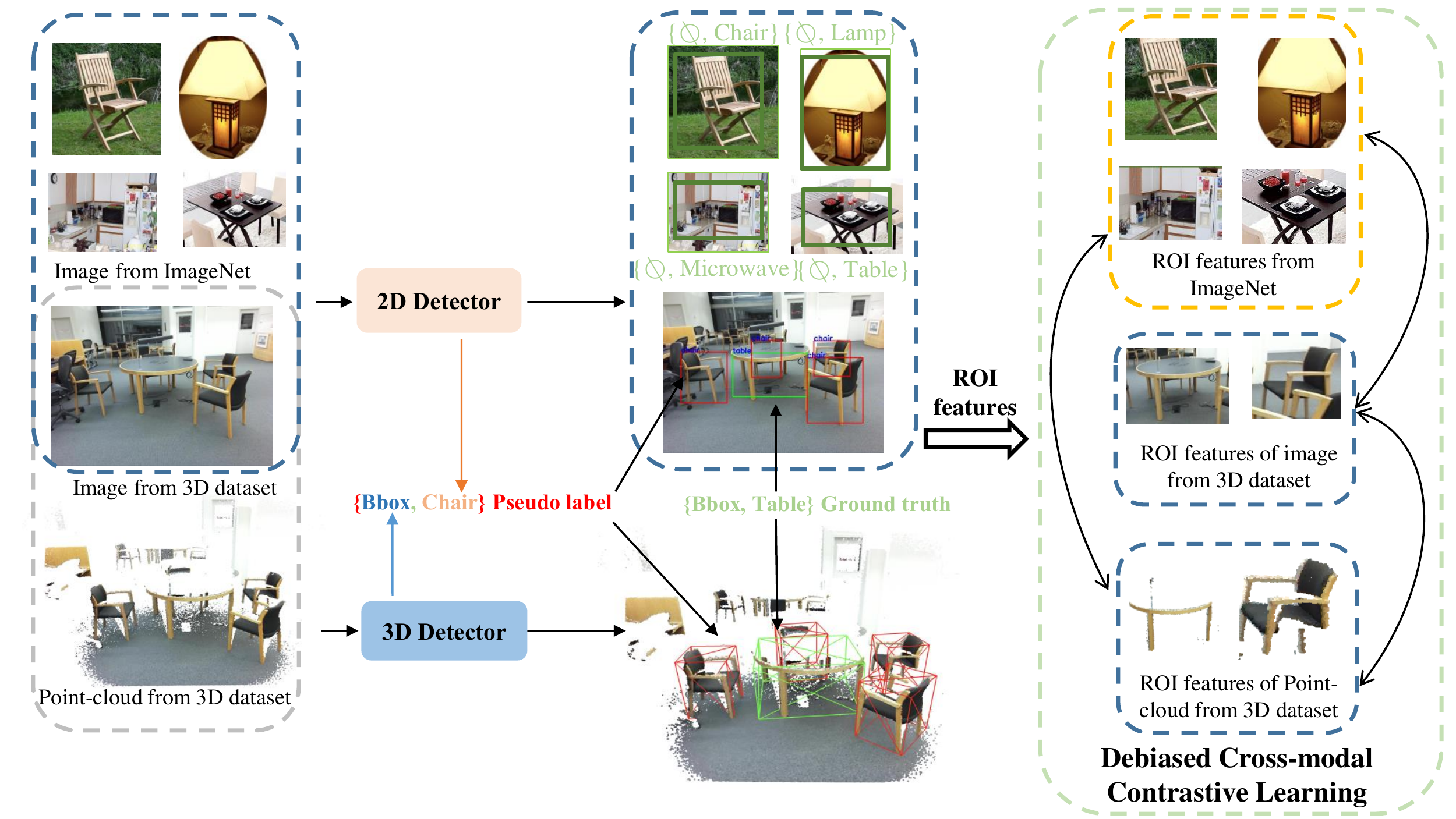}
    \caption{Overview of the phase 2 of OV-3DETIC, which is the core and a synergy of two components: 1) We take advantage of two modalities, image modality for classification and point-cloud modality for localization, to generate pseudo labels for unseen classes, and 2) we design a debiased cross-modal contrastive learning to transfer the knowledge from the image modality to the point-cloud modality. Note that before phase 2, we first train both the 2D detector and the 3D detector as similar to Detic \cite{zhou2022detecting}. The green brackets denote ground truth of seen classes. For ImageNet1K, the "bbox" is not provided, so denoted as $\varnothing$. The red brackets denote pseudo label, where bounding boxes come from the output of the 3D detector, and classes come from the output of the 2D detector. 
    }
    \label{fig:pseudolabel}
\end{figure*}

\section{Method}
\label{method}

\subsection{Notation and Preliminaries}
We use $\mathbf{I} \in \mathcal{R}^{3 \times H \times W}$ to represent image, and $\mathbf{P} = \{\mathbf{p}_i \in \mathcal{R}^{3}, i = 1, 2, 3 ..., N\}$ to represent point-cloud, where $N$ is the point number in the point-cloud. During training, we use 1) point-cloud dataset denoted as $\mathcal{D}^{pc} = \{(\mathbf{P},  {(\mathbf{b}_{3D} \in \mathcal{R}^7, \mathbf{c}_{3D})_k})_j\}_{j=1}^{|\mathcal{D}^{pc}|}$, with vocabulary size $\mathcal{C}_{pc}$, where $\mathbf{b}_{3D}$ is the annotation of 3D bounding box, $\mathbf{c}_{3D}$ is the corresponding classification label; 2) paired image dataset denoted as $\mathcal{D}^{img} = \{\mathbf{I}_j\}_{j=1}^{|\mathcal{D}^{img}|}$; 3) ImageNet1K dataset denoted as $\mathcal{D}^{ign} = \{(\mathbf{I}, \mathbf{c}^{ign}_{2D})_j\}_{j=1}^{|\mathcal{D}^{ign}|}$, with vocabulary $\mathcal{C}_{ign}$, where $\mathbf{c}^{ign}_{2D}$ denotes classfication label of the image in ImageNet1K. During test, we evaluate on the vocabulary $ \mathcal{C}_{test}$, where $\mathcal{C}_{ign} \geq \mathcal{C}_{test} \textgreater \mathcal{C}_{pc}$. 

A typical point-cloud detector deals with localization and classification, where the localization module outputs bounding boxes $\hat{\mathbf{b}}_{3D} \in \mathcal{R}^{7}$ and we could get its corresponding point-cloud ROI features $\mathbf{f}_{3D}$. Similarly, we can project the 3D bounding box into 2D image via projection matrix $K$, i.e., $\hat{\mathbf{b}}_{2D} \in \mathcal{R}^{4} $ and index the corresponding image ROI feature  $\mathbf{f}_{2D}$. Then we would use the $\mathbf{f}_{3D}$ and $\mathbf{f}_{2D}$ to predict the class of the objects in bounding boxes.

\subsection{OV-3DETIC: Open-Vocabulary 3D Detector with Image Classes}
\label{sec:ov-3detic}
We use ImageNet1K to broaden the vocabulary of the point-cloud detector. To transfer the knowledge contained in ImageNet1K to the point-cloud detector, we introduce images paired to point-clouds to bridge them, but do not use any extra labels. We design a two-phase training strategy to enable open-vocabulary point-cloud detection. The first-phase training is similar to Detic \cite{zhou2022detecting}, which aims at leveraging ImageNet to help the 2D detector be able to learn sufficient knowledge. The second-phase is the core of the proposed OV-3DETIC, which aims at transferring the knowledge of the 2D detector to the 3D detector via customized pseudo-label strategy and debiased cross-modal contrastive learning, as shown in Figure. \ref{fig:pseudolabel}. During inference, we only need the 3D detector without any extra models or modalities. OV-3DETIC is developed on top of 3DETR \cite{misra2021end}, and we use DETR \cite{carion2020end} as the 2D detector. The details are illustrated below.

In the first stage, the point-cloud $\mathbf{P}$ in $\mathcal{D}^{pc}$ is input to the 3D detector, which is supervised by the ground truth of $\{\mathbf{b}_{3D}, \mathbf{c}_{3D}\}$ for seen objects. The paired images $\mathbf{I}$ from $\mathcal{D}^{img}$ are input into the 2D detector, which is supervised by projected 3D bounding boxes and the corresponding class, denoted as $\{\mathbf{b}_{3D} \times K, \mathbf{c}_{3D}\}$. It is noteworthy to mention that the images $\mathbf{I}$ from from ImageNet1K $\mathcal{D}^{ign}$ are also input into the same 2D detector although only image-level labels (classification labels) are provided. Following Detic \cite{zhou2022detecting}, we choose the max-size proposal $\mathbf{f}_{max\-size}$ and apply classification label $ \mathbf{c}^{ign}_{2D}$ to supervised the classifier in the 2D detector. The loss in the phase 1 training is given by
\begin{equation}
\begin{split}
L^{phase1} =& L_{box}^{3D}(\mathbf{b}_{3D}, \hat{\mathbf{b}}_{3D}) + L_{cls}^{3D}(\mathbf{c}_{3D}, W_{3D}\mathbf{f}_{3D}) + \\
&L_{box}^{2D}(\mathbf{b}_{3D} \times K, \hat{\mathbf{b}}_{2D})+ L_{cls}^{2D}(\mathbf{c}_{3D}, W_{2D}\mathbf{f}_{2D}) + L_{cls}^{ign}(\mathbf{c}_{2D}^{ign}, W_{2D}\mathbf{f}_{max\-size}),
\end{split}
\end{equation}
where $L_{box}^{3D}$ and $L_{box}^{2D}$ follow \cite{misra2021end} and \cite{carion2020end}. $L_{cls}^{3D}$ and $L_{cls}^{2D}$ are cross-entropy loss for classification. $W_{3D}$ and $W_{2D}$ denote the classifier in the 3D detector and the 2D detector, respectively.

In phase 2, we first generate pseudo labels for unseen classes. The pseudo labels contain two parts: bounding box from the 3D detector, and class from the 2D detector. Specifically, since we already exploit ImageNet1K to train the 2D detector in the phase 1, as similar to Detic \cite{zhou2022detecting}, the 2D detector is able to classify unseen classes so that we can use its classification results as relatively accurate pseudo labels. 
To leverage this, we crop the image region of the projected 3D detection and use the 2D detector's classifier to generate its class label.
For the bounding box pseudo labels, we take advantage of the generalizability of localization for the point-cloud detector, as mentioned in Figure. \ref{fig:localization_veryfy}. Besides using pseudo labels, we also design a debiased cross-modal contrastive learning to better transfer the knowledge from 2D modality to point-cloud modality. 
Note that there is significant synergy between this pseudo-label strategy and our proposed debiased cross-modal contrastive learning. The pseudo labels help address false negatives, cross-modal contrastive learning identifies these new detections and allows both modalities to learn from them, in turn generating better pseudo-labels.
Thus the pseudo labels are iteratively updated to be of higher quality. Overall, the total loss in phase 2 is given by

\begin{equation}
\begin{split}
L^{phase2} =& L_{box}^{3D}(\mathbf{\bar{b}}_{3D}, \hat{\mathbf{b}}_{3D}) + L_{cls}^{3D}(\mathbf{\bar{c}}_{3D}, W_{3D}\mathbf{f}_{3D}) + 
L_{box}^{2D}(\mathbf{\bar{b}}_{3D} \times K, \hat{\mathbf{b}}_{2D})+\\ &L_{cls}^{2D}(\mathbf{\bar{c}}_{3D}, W_{2D}\mathbf{f}_{2D}) + L_{cls}^{ign}(\mathbf{c}_{2D}^{ign}, W_{2D}\mathbf{f}_{max\-size}) + L_{DECC},
\end{split}
\end{equation}

where $\mathbf{\bar{b}}_{3D}$, $\mathbf{\bar{b}}_{2D}$, and $\mathbf{\bar{c}}_{3D}$ come from either ground truth or pseudo labels. Next, we will illustrate the detail of debiased cross-modal contrastive loss $L_{DECC}$.


\begin{figure*}[!t]
    \centering
    \includegraphics[width=.9\textwidth]{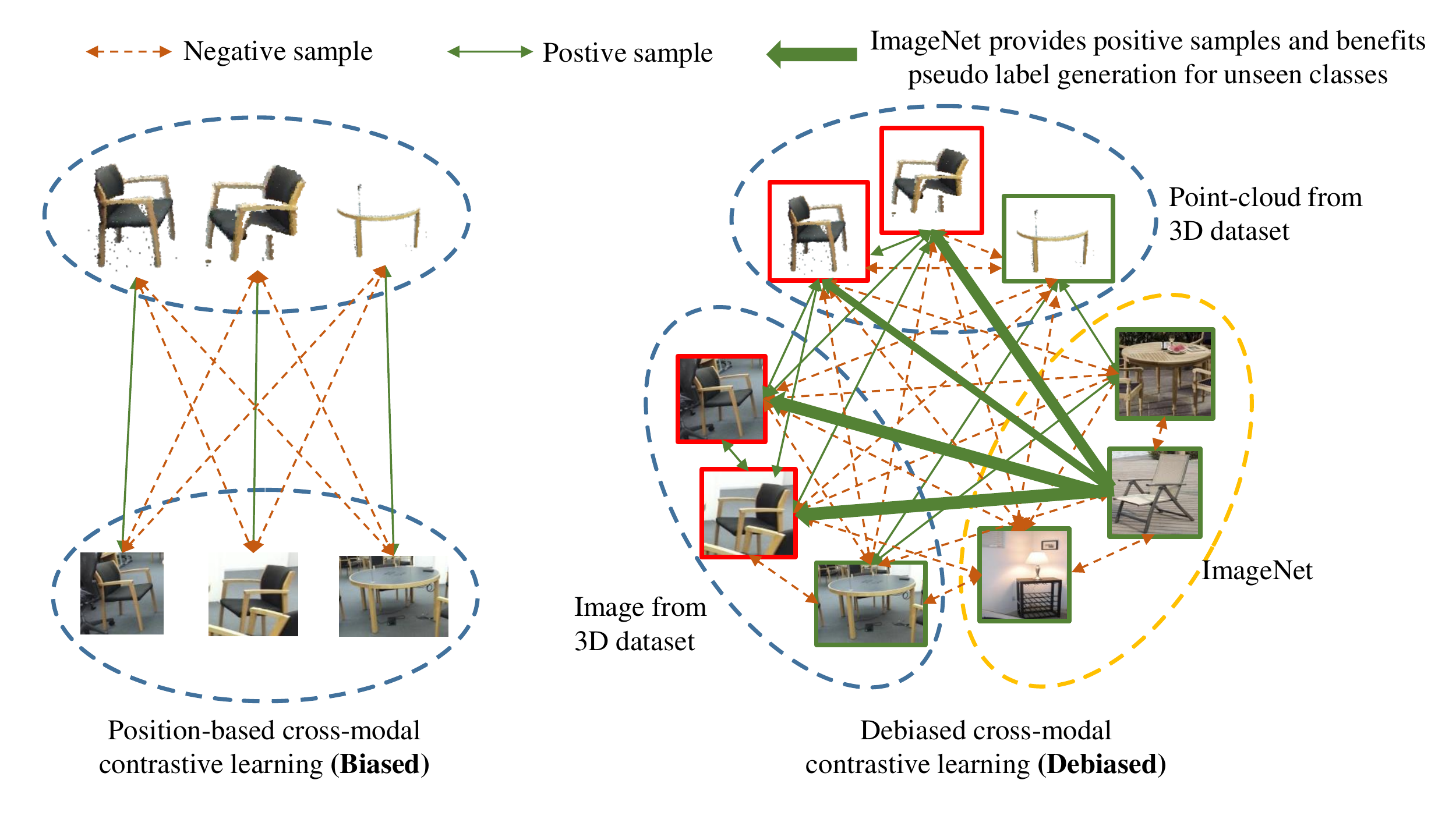}
    \caption{Typical biased cross-modal contrastive learning (left) and the proposed debiased cross-modal contrastive learning (right). Position-based cross-modal contrastive learning follows the position correspondence with one-to-one match. This may result in assigning inaccurate negative sample. The proposed debiased cross-contrastive learning (DECC) leverages pseudo labels to facilitate the biased issues. The red outline denotes it is an unseen class and we assign a pseudo label to it, and the green outline denotes the seen class with ground truth. }
    \label{fig:debiased}
\end{figure*}

\subsection{Debiased Cross-modal Contrastive Learning}
\label{sec:decc}
Before we in-depth illustrate the proposed debiased cross-modal contrastive learning, we first discuss the typical biased contrastive learning in the following. Cross-modal contrastive learning has been demonstrated as a powerful method used for knowledge transferring \cite{jiang2019transferable,radford2021learning,minderer2022simple}, and enabling zero-shot/open vocabulary classification and detection.
Using instance discrimination as the pretext task, typical contrastive learning \cite{radford2021learning,chen2020simple,he2020momentum} pulls the semantically-close neighbors together and pushes away non-neighbors. In previous works \cite{radford2021learning,minderer2022simple}, dissimilar (negative) points are typically taken to be randomly sampled datapoints, implicitly accepting that these points may actually express the same underlying concepts. For example, CLIP \cite{radford2021learning} conducts contrastive learning on coupled text and image, while the text caption may describe a bunch of images that have similar semantics. This issue is studied as noisy matching \cite{wu2021data} or biased contrastive learning \cite{chuang2020debiased}. In the image-point-cloud field, existing studies mainly focused on matching the features from the same 3D positions \cite{hou2021pri3d,liu2021learning,park2022detmatch}. This results in biases, as shown in left part of Figure. \ref{fig:debiased}. We can observe that both within and across modalities, some objects with the same classes are inaccurately assigned as negative samples.

In order to mitigate the biased issue in contrastive learning, we propose a debiased cross-modal contrastive learning (DECC), as shown in right part of Figure. \ref{fig:debiased}. We take the ROI features $\mathbf{f}_{3D}$ from the 3D detector and  $\mathbf{f}_{2D}$ from the 2D detector, and use the ground truth or pseudo labels $\mathbf{\bar{c}}_{3D}$ to assign the positive and negative samples. For a mini-batch of $\mathbf{f}_{3D}$ and $\mathbf{f}_{2D}$ with total $M$ features from two modalities, we exploit a linear projection and normalize them as similar as CLIP \cite{radford2021learning}, denoting their hidden representations as $h_i$, where $i=1,2,3,...,M$. Then the loss is given by
\begin{equation}
     L_{DECC}= -\frac{1}{M}\sum_{i=1}^M\log\frac{\sum_{t=0}^me^{h_i^\top  h_t/\tau(dist_{it})}}{\sum_{j=0}^Me^{h_i^\top h_j/\tau_0}}, 
\end{equation}

where $m$ is the number of positive sample corresponding to $h_i$ ($m <=M$), $\tau_0$ is the base temperature. It is noteworthy to mention that different from previous contrastive learning with constant temperature $\tau$ for each sample, we here adopt a distance-aware temperature $\tau(dist_{ij}) = \tau_0 \times \gamma^{dist_{ij}}$ for positive sample, where $\gamma$ are hyperparameters. In particular, we calculate the Euclidean distance of the two samples in 3D space. Note that if the one of the sample from ImageNet, we consider the distance between ImageNet and any of other sample as 1. The distance-aware temperature aims at facilitating the uncertain pseudo labeling. It is prone to confuse the classifier for the two samples distributing distantly, thus we scale the temperature with distance-aware strategy to modify it. If we set $\gamma$ to 1.0, DECC degenerates to debiased cross-modal contrastive learning with constant temperature, and if we set $\gamma$ to +$\infty$, DECC degenerates to position-based cross-modal contrastive learning.
\label{sec:MethodDECC}

\section{Experiments}
In this section, we compare the proposed OV-3DETIC with popular baselines on two widely used 3D detection datasets, SUN RGB-D \cite{song2015sun} and ScanNet\cite{dai2017scannet}. Then we conduct sufficient analysis and ablation studies to explore why OV-3DETIC works. The details are illustrated in the below.

\subsection{Datasets and Evaluation Metric}
\textbf{SUN RGB-D} \cite{song2015sun} and \textbf{ScanNet} \cite{dai2017scannet} are two widely used 3D detection datasets. We follow the data preprocessing procedure from VoteNet \cite{qi2019deep}, except for the chosen of unseen classes. We randomly select 10 unseen classes for both two datasets. Specifically, {toilet, bed, chair, bathtub, sofa, dresser, scanner, fridge, lamp} and {desk} are unseen classes for SUN RGB-D, while {table, night stand, cabinet, counter, garbage bin, bookshelf, pillow, microwave, sink} and {stool} are seen classes. {toilet, bed, chair, sofa, dresser, table, cabinet, bookshelf} and {pillow} are unseen classes for ScanNet, and {bathtub, fridge, desk, night stand, counter, door, curtain, box, lamp} and {bag} are seen classes. The metrics we use in the experiments are Average Precision (AP), mean Average Precision (mAP) and Average Recall (AR) at IoU thresholds of 0.25, denoted as $AP_{25}$, $mAP_{25}$, $AR_{25}$, respectively.

\subsection{Experimental Setup and Implementation Details}

\paragraph{Baseline and Implementation}
The proposed OV-3DETIC is mainly based on 3DETR \cite{misra2021end}, yet it can also generalize to other point-cloud detectors. We regard the point-cloud detectors that are trained on seen classes as our baselines. During training, a minibatch of data consists of the point cloud with its paired image and several images from ImageNet. 3DETR as the 3D detector consumes the point cloud, and DETR \cite{carion2020end} as the 2D detector deals with images. Note that only classification loss is used in the DETR when the input is images from ImageNet. We select the max-size ROI features detected by DETR and apply the classification loss on it, as following Detic \cite{zhou2022detecting}.
$\tau_{0}$ and $\gamma$ are two hyperparameters in the distance-aware temperature strategy. In practice, $\tau_{0}$ is set to 0.2, $\gamma$ is set to 1.1, and the distance between the image from ImageNet and any other sample is 1.0. For both 3DETR and DETR, a unified learning rate is set to $2\times10^{-5}$, with batch size of 4 for each GPU, and we train our model on 8 RTX 2080 TI GPUs. We train 200 epochs for phase 1, and 200 epochs for phase 2. 


\paragraph{Pseudo Label Generation} 
Pseudo label generation includes two components: 3D bounding box and category. We first generate initial pseudo labels after finishing phase 1, and update them iteratively. As we discussed in Section \ref{sec:ov-3detic}, the 3D detector performs generalizable object localization, while the 2D detector is trained on the ImageNet dataset with sufficient knowledge. Therefore, the bounding box comes from the 3D detector and the class label comes from the 2D detector. Furthermore, we only keep proposals with high confidence and filter out the proposals that are duplicated with ground truth of seen classes or no point in it. Finally, only top-$k$ proposals are kept for each unseen class, and $k$ increases linearly every 50 epochs in phase 2. $k$ is default as 50, and increases 10 (default) every 50 epochs.

\subsection{Main Results}

\begin{table}[H]
  \caption{Detection results ($AP_{25}$) on unseen classes of SUN RGB-D.}
  \centering
  \resizebox{1\textwidth}{!}{
      \begin{tabular}{cccccccccccc}
        \toprule
        Method  &{{toilet}}  &{{bed}}  &{{chair}}  &{{bathtab}}  &{{sofa}}  &{{dresser}}  &{{scanner}}  &{{fridge}}  &{{lamp}}  &{{desk}}  &{{mean}} \\
        \midrule
        GroupFree3D \cite{Liu2021GroupFree3O}&0.23		&0.04		&1.25		    &0.03		&0.21		&0.21		&0.14		    &0.10		&0.03		&3.02		&0.53		\\
        VoteNet \cite{Qi2019DeepHV}  		&0.12		&0.05		&1.12		    &0.03		&0.09		&0.15		&0.06		    &0.11		&0.04		&2.10		&0.39		\\
        H3DNet \cite{Zhang2020H3DNet3O}		&0.24		&0.10		&1.28	&0.05		&0.22		&0.22		&0.13	    	&0.14		&0.03		&6.09		&0.85		\\
        3DETR \cite{misra2021end}   		&1.57		&0.23		&0.77		    &0.24		&0.04		&0.61		&\textbf{0.32}	&0.36		&0.01		&8.92		&1.31\\
        \midrule
        \midrule
        OV-PointCLIP \cite{zhang2021pointclip}      &7.90		&2.84		&\textbf{3.28}		    &0.14		&1.18		&0.39		&0.14		    &0.98		&\textbf{0.31}		&5.46		&2.26		\\
        OV-Image2Point \cite{xu2021image2point}     &2.14		&0.09		&3.25		    &0.01		&0.15		&0.55		&0.04		    &0.27		&0.02		&5.48		&1.20		\\
        Detic-ModelNet \cite{zhou2022detecting}	    &3.56		&1.25		&2.98       	&0.02		&1.02		&0.42		&0.03	    	&0.63		&0.12		&5.13		&1.52		\\
        Detic-ImageNet \cite{zhou2022detecting}	    &0.01		&0.02		&0.19		    &0.00		&0.00		&1.19		&0.23		    &0.19		&0.00		&7.23		&0.91       \\
        \midrule
        \midrule
        Ours            &\textbf{43.97}	&\textbf{6.17}	&0.89	&\textbf{45.75}	&\textbf{2.26}	&\textbf{8.22}	&0.02	&\textbf{8.32}	&0.07	&\textbf{14.60}	&\textbf{13.03}		\\
        Improvement                         &\textcolor{ForestGreen}{\textbf{+36.07}}	&\textcolor{ForestGreen}{\textbf{+3.33}}	&\textcolor{ForestGreen}{\textbf{-2.39}}	&\textcolor{ForestGreen}{\textbf{+45.51}}	&\textcolor{ForestGreen}{\textbf{+1.08}}	&\textcolor{ForestGreen}{\textbf{+7.03}}	&\textcolor{ForestGreen}{\textbf{-0.30}}	&\textcolor{ForestGreen}{\textbf{+7.34}}	&\textcolor{ForestGreen}{\textbf{-0.24}}	&\textcolor{ForestGreen}{\textbf{+5.68}}	&\textcolor{ForestGreen}{\textbf{+10.77}}		\\
        \bottomrule
      \end{tabular}
  }
  \label{tab:SOTARGBD}
\end{table}

\begin{table}[H]
  \caption{Detection results ($AP_{25}$) on unseen classes of ScanNet.}
  \centering
  \resizebox{1\textwidth}{!}{
      \begin{tabular}{cccccccccccc}
        \toprule
        Method  &{{toilet}}  &{{bed}}  &{{chair}}  &{{sofa}}  &{{dresser}}  &{{table}}  &{{cabinet}}  &{{bookshelf}}  &{{pillow}}  &{{sink}}  &{{mean}} \\
        \midrule
        GroupFree3D \cite{Liu2021GroupFree3O} 	&0.63 	&0.52 	&1.52 	&0.52 	&0.20 	&0.59 	&0.52 	&0.25 	&0.01 	&0.15 	&0.49 \\
     	VoteNet \cite{Qi2019DeepHV}             &0.04 	&0.02 	&0.12 	&0.00 	&0.02 	&0.11 	&0.07 	&0.05 	&0.00 	&0.00 	&0.04 \\
     	H3DNet \cite{Zhang2020H3DNet3O} 	    &0.55 	&0.29 	&1.70 	&0.31 	&0.18 	&0.76 	&0.49 	&0.40 	&0.01 	&0.10 	&0.48 \\
     	3DETR \cite{misra2021end}    	        &2.60 	&0.81 	&0.90 	&1.27 	&0.36 	&1.37 	&0.99 	&2.25 	&0.00 	&0.59 	&1.11 \\
        \midrule
        \midrule
        OV-PointCLIP \cite{zhang2021pointclip}      &6.55		&2.29		&6.31		    &3.88		&0.66		&7.17		&0.68		    &2.05		&0.55		&0.79		&3.09		\\
        OV-Image2Point \cite{xu2021image2point}     &0.24		&0.77		&0.96		    &1.39		&0.24		&2.82		&0.95		    &0.91		&0.00		&0.08		&0.84		\\
        Detic-ModelNet \cite{zhou2022detecting}	    &4.25		&0.98		&4.56       	&1.20		&0.21		&3.21		&0.56	    	&1.25		&0.00		&0.65		&1.69		\\
        Detic-ImageNet \cite{zhou2022detecting} 	&0.04	    &0.01   	&0.16   	    &0.01   	&0.52	    &1.79   	&0.54   	    &0.28   	&0.04	    &0.70       &0.41       \\
        \midrule
        \midrule
        Ours            &\textbf{48.99}	&\textbf{2.63}	&\textbf{7.27}	&\textbf{18.64}	&\textbf{2.77}	&\textbf{14.34}	&\textbf{2.35}	&\textbf{4.54}	&\textbf{3.93}	&\textbf{21.08}	&\textbf{12.65}  \\
        Improvement      			  &\textcolor{ForestGreen}{\textbf{+42.44}}  &\textcolor{ForestGreen}{\textbf{+0.34}}  &\textcolor{ForestGreen}{\textbf{+0.96}}  &\textcolor{ForestGreen}{\textbf{+14.76}}  &\textcolor{ForestGreen}{\textbf{+2.11}}  &\textcolor{ForestGreen}{\textbf{+7.17}}  &\textcolor{ForestGreen}{\textbf{+1.36}}  &\textcolor{ForestGreen}{\textbf{+2.29}}  &\textcolor{ForestGreen}{\textbf{+3.38}}  &\textcolor{ForestGreen}{\textbf{+20.29}}  &\textcolor{ForestGreen}{\textbf{+9.56}}       \\
        \bottomrule
      \end{tabular}
  }
  \label{tab:SOTASCANNET}
\end{table}

As there is no baseline directly solving the problem of open-vocabulary 3D point cloud detection, we mainly compare OV-3DETIC with state-of-the-art 3D point cloud detectors \cite{Liu2021GroupFree3O,Qi2019DeepHV,Zhang2020H3DNet3O,misra2021end} and some well-known works \cite{zhang2021pointclip,xu2021image2point,zhou2022detecting} that study either transferability in point-cloud or 2D open-vocabulary detection.
Specifically, the baselines we use include:
\begin{itemize}
    \item GroupFree3D \cite{Liu2021GroupFree3O}, VoteNet \cite{Qi2019DeepHV}, H3DNet \cite{Zhang2020H3DNet3O}, 3DETR \cite{misra2021end} are well-known and representative 3D point cloud detectors that are chosen as our baselines. Specifically, these four baselines are trained on the seen classes while being tested on the unseen.
    
    \item The second is PointCLIP \cite{zhang2021pointclip} which bridges the point-cloud and text domain. We use it directly as a pre-trained open vocabulary 3D classifier and replace the classifier of 3DETR with PointCLIP. This baseline is denoted as \textbf{OV-PointCLIP}, which is similar to well-known 2D open-vocabulary detection works \cite{bansal2018zero,gu2021zero,zhou2022detecting} that replace the classifier in the detector with the generalizable classifier. 
    
    \item Besides, Xu et al.\cite{xu2021image2point} transfer the image pre-trained transformer to the point cloud by copying or inflating the weights. Similarly, we copy the weights of the transformer and the classifier from pre-trained DETR (pre-trained on COCO \cite{lin2014microsoft}) to 3DETR and finetune the set aggregation module and the 3D box header. We term this baseline as \textbf{OV-Image2Point}.
    
    \item Moreover, Detic \cite{zhou2022detecting} leverages large-scale classification dataset (ImageNet) to broaden the 2D detector, here we directly extend the idea to 3D open-vocabulary detection. Specifically, we consider two manners, extend the classifier via ModelNet or ImageNet, and term them as \textbf{Detic-ModelNet} and \textbf{Detic-ImageNet}, respectively. 
\end{itemize}

The results are presented in Table. \ref{tab:SOTARGBD} and Table. \ref{tab:SOTASCANNET}. We can observe that \textbf{OV-PointCLIP} achieves $mAP_{25}$ of 2.26\% and 3.09\%, outperforming the other baselines on both SUN RGB-D and ScanNet. Furthermore, \textbf{Detic-ModelNet} reaches $mAP_{25}$ of 1.52\% and 1.69\% on SUN RGB-D and ScanNet, respectively. It can be observed that both \textbf{OV-PointCLIP} and \textbf{Detic-ModelNet} outperforms \textbf{OV-Image2Point} and \textbf{Detic-ImageNet}. Indeed, \textbf{OV-PointCLIP} and \textbf{Detic-ModelNet} enhances the classifier in the detector via introducing the pretraining on ModelNet, which is a 3D classification dataset, while \textbf{OV-Image2Point} and \textbf{Detic-ImageNet} try to transfer knowledge from image (COCO and ImageNet) to point cloud. This contrast shows that the modality gap between 2D and 3D does indeed hinder knowledge transfer. The results also demonstrate that directly plugging and playing the Detic method on 3DETR with introducing ImageNet is infeasible to transferring the knowledge from 2D image to 3D, which is also mentioned in the discussion in related work. Nonetheless, our method achieves $mAP_{25}$ of 13.03\% and 12.65\% on SUN RGB-D and ScanNet, respectively, which proves the proposed OV-3DETIC can indeed make use of the image knowledge, to achieve open-vocabulary 3D detection.

\label{sec:MainResult}

\subsection{Results on Other Randomly Resampled Unseen Classes}
In order to verify the robustness against the sampling of unseen classes, we further shuffle and randomly resample multiple sets of unseen classes. As shown in Table. \ref{tab:SOTARGBD_resample_set_1}, \ref{tab:SOTARGBD_resample_set_2}, \ref{tab:SOTARGBD_resample_set_3}, for different setting of unseen classes, we compare OV-3DETIC with the baseline method (3DETR that trained on seen classes). The results show that OV-3DETIC outperforms the baseline and achieves 13.3\% $mAP_{25}$ that averages over all sets. Besides, even the worst-performing set (Set 2) overtakes the baseline by a large margin of 11.59\%, which demonstrates OV-3DETIC is robust to the sampling of unseen classes. By the way, we observe that there are some classes that perform well across all sets of unseen classes, such as "toilet" and "bathtub", while there are also some cases that perform poorly, such as "microwave" and "lamp". This phenomenon shows that the difficulty of each class does indeed affect the final performance.

\begin{table}
  \caption{Supplementary results ($AP_{25}$) on resampled unseen classes of SUN RGB-D. (Set 1)}
  \centering
  \resizebox{1\textwidth}{!}{
      \begin{tabular}{cccccccccccc}
        \toprule
        Method  &{{bed}}  &{{bathtub}}  &{{dresser}}  &{{fridge}}  &{{desk}}  &{{stand}}  &{{counter}}  &{{bookshelf}}  &{{microwave}}  &{{stool}}  &{{mean}}  \\
        \midrule
        3DETR \cite{misra2021end}   		&0.20  &0.02  &1.24  &0.17  &2.68  &0.18  &0.29  &0.19  &0.00  &0.08  &0.51  \\
        \midrule
        \midrule
        Ours                                &\textbf{1.35}	&\textbf{47.25}	&\textbf{9.51}	&\textbf{10.17}	&\textbf{13.39}	&\textbf{25.96}	&\textbf{3.65}	&\textbf{17.28}	&\textbf{0.71}	&\textbf{4.58}	&\textbf{13.39}  \\
        
        Improvement                         &\textcolor{ForestGreen}{\textbf{+1.15}}  &\textcolor{ForestGreen}{\textbf{+47.23}}  &\textcolor{ForestGreen}{\textbf{+8.27}}  &\textcolor{ForestGreen}{\textbf{+10.00}}  &\textcolor{ForestGreen}{\textbf{+10.71}}  &\textcolor{ForestGreen}{\textbf{+25.78}}  &\textcolor{ForestGreen}{\textbf{+3.36}}  &\textcolor{ForestGreen}{\textbf{+17.09}}  &\textcolor{ForestGreen}{\textbf{+0.71}}  &\textcolor{ForestGreen}{\textbf{+4.5}}  &\textcolor{ForestGreen}{\textbf{+12.88}}    \\
        
        \bottomrule
      \end{tabular}
  }
  \label{tab:SOTARGBD_resample_set_1}
\end{table}

\begin{table}
  \caption{Supplementary results ($AP_{25}$) on resampled unseen classes of SUN RGB-D. (Set 2)}
  \centering
  \resizebox{1\textwidth}{!}{
      \begin{tabular}{cccccccccccc}
        \toprule
        Method  &{{chair}}  &{{bathtub}}  &{{sofa}}  &{{lamp}}  &{{desk}}  &{{table}}  &{{counter}}  &{{pillow}}  &{{sink}}  &{{stool}}  &{{mean}}  \\
        \midrule
        3DETR \cite{misra2021end}   		&1.12  &0.02  &0.22  &0.00  &0.37  &0.19  &0.46  &0.00  &0.18  &0.04  &0.26\\
        \midrule
        \midrule
        Ours                                &\textbf{4.38}  &\textbf{45.70}  &\textbf{3.23}  &\textbf{2.72}  &\textbf{7.62}  &\textbf{10.64}  &\textbf{10.39}  &\textbf{4.16}  &\textbf{27.45}  &\textbf{2.16}  &\textbf{11.85}		\\
        Improvement                         &\textcolor{ForestGreen}{\textbf{+3.26}}  &\textcolor{ForestGreen}{\textbf{+45.68}}  &\textcolor{ForestGreen}{\textbf{+3.01}}  &\textcolor{ForestGreen}{\textbf{+2.72}}  &\textcolor{ForestGreen}{\textbf{+7.25}}  &\textcolor{ForestGreen}{\textbf{+10.45}}  &\textcolor{ForestGreen}{\textbf{+9.93}}  &\textcolor{ForestGreen}{\textbf{+4.16}}  &\textcolor{ForestGreen}{\textbf{+27.27}}  &\textcolor{ForestGreen}{\textbf{+2.12}}  &\textcolor{ForestGreen}{\textbf{+11.59}}		\\
        
        \bottomrule
      \end{tabular}
  }
  \label{tab:SOTARGBD_resample_set_2}
\end{table}

\begin{table}
  \caption{Supplementary results ($AP_{25}$) on resampled unseen classes of SUN RGB-D. (Set 3)}
  \centering
  \resizebox{1\textwidth}{!}{
      \begin{tabular}{cccccccccccc}
        \toprule
        Method  &{{toilet}}  &{{bathtub}}  &{{sofa}}  &{{fridge}}  &{{lamp}}  &{{table}}  &{{counter}}  &{{bin}}  &{{microwave}}  &{{stool}}  &{{mean}}  \\
        \midrule
        3DETR \cite{misra2021end}   		&2.72  &0.05  &0.29  &0.29  &0.02  &1.37  &0.39  &0.42  &0.00  &0.07  &0.56\\
        \midrule
        \midrule
        Ours                                &\textbf{50.97}  &\textbf{44.30}  &\textbf{4.73}  &\textbf{13.60}  &\textbf{0.01}  &\textbf{10.27}  &\textbf{4.34}  &\textbf{16.11}  &\textbf{0.72}  &\textbf{2.54}  &\textbf{14.76}		\\
        Improvement                         &\textcolor{ForestGreen}{\textbf{+48.25}}  &\textcolor{ForestGreen}{\textbf{+44.25}}  &\textcolor{ForestGreen}{\textbf{+4.44}}  &\textcolor{ForestGreen}{\textbf{+13.31}}  &\textcolor{ForestGreen}{\textbf{-0.01}}  &\textcolor{ForestGreen}{\textbf{+8.90}}  &\textcolor{ForestGreen}{\textbf{+3.95}}  &\textcolor{ForestGreen}{\textbf{+15.69}}  &\textcolor{ForestGreen}{\textbf{+0.72}}  &\textcolor{ForestGreen}{\textbf{+2.47}}  &\textcolor{ForestGreen}{\textbf{+14.20}}    \\
        
        \bottomrule
      \end{tabular}
  }
  \label{tab:SOTARGBD_resample_set_3}
\end{table}

\begin{table}
  \caption{Ablation study on different components.}
  \centering
  \resizebox{1\textwidth}{!}{
      \begin{tabular}{cccccccc}
        \toprule
        \multirow{2}*{Baseline}    &\multirow{2}*{Pseudo Label}   &Augmentation        &Position        &Class               &Distance-Aware         &\multirow{2}*{$mAP_{25}$}     &\multirow{2}*{$AR_{25}$} \\
        ~                          &~                             &Based CL                       &Based CL                   &Based CL                       &Temperature            &~                             &~                       \\
        
        \midrule
        \checkmark  &   &   &   &   &   &1.31   &7.00    \\
        \midrule
        \midrule
        \checkmark  &\checkmark   &             &           &             &             &12.55           &38.35           \\
        \checkmark  &\checkmark   &\checkmark   &           &             &             &10.48           &36.44           \\
        \checkmark  &\checkmark   &             &\checkmark &             &             &12.36           &37.35           \\
        \checkmark  &\checkmark   &             &           &\checkmark   &             &12.92           &\textbf{42.48}  \\
        \checkmark  &\checkmark   &             &           &\checkmark   &\checkmark   &\textbf{13.03}  &37.71           \\
        
        \bottomrule
      \end{tabular}
  }
  \label{tab:ABLATION}
\end{table}

\subsection{Analysis and Ablation Study}
\paragraph{Ablation Study on Different Components}
We conduct ablation studies on SUN RGB-D. The results of unseen classes are reported in Table \ref{tab:ABLATION}. Our baseline is 3DETR, trained only on seen categories. "Pseudo Label" denotes to train with the pseudo label. "Augmentation Based CL" follows SimCLR \cite{chen2020simple,he2020momentum} that utilize data augmentation strategy in contrastive learning, which, however, is a biased contrastive learning setting. Position-based contrastive learning only takes the same instance from the paired image / point-cloud as positive. Class-based contrastive learning takes positive so long as two samples belong to the same category. "Distance-Aware Temperature" refers to the strategy that is introduced in Section \ref{sec:MethodDECC}. We can observe that, first, the pseudo label brings the largest improvement, actually, the pseudo label implicitly transfers the knowledge contained in ImageNet to a different modality, the point-cloud modality, via providing useful class labels. This not only validates our hypothesis that the 3D detector functions as a proposal network with strong generalization ability, but also demonstrates the effectiveness of introducing large-scale image-level supervision for classification. Second, class-based contrastive learning and distance-aware temperature further significantly improve the performance, while traditional augmentation-based contrastive learning and position-based contrastive learning hurt the performance, which indicates the weakness of biased contrastive learning, and demonstrates the proposed debiased cross-modal contrastive learning indeed benefits the point-cloud detector to learn general representations. 

\begin{figure*}[htbp]
    \centering
    \resizebox{1\textwidth}{!}{
        \subfigure[Baseline]{
            \includegraphics[width=0.25\textwidth]{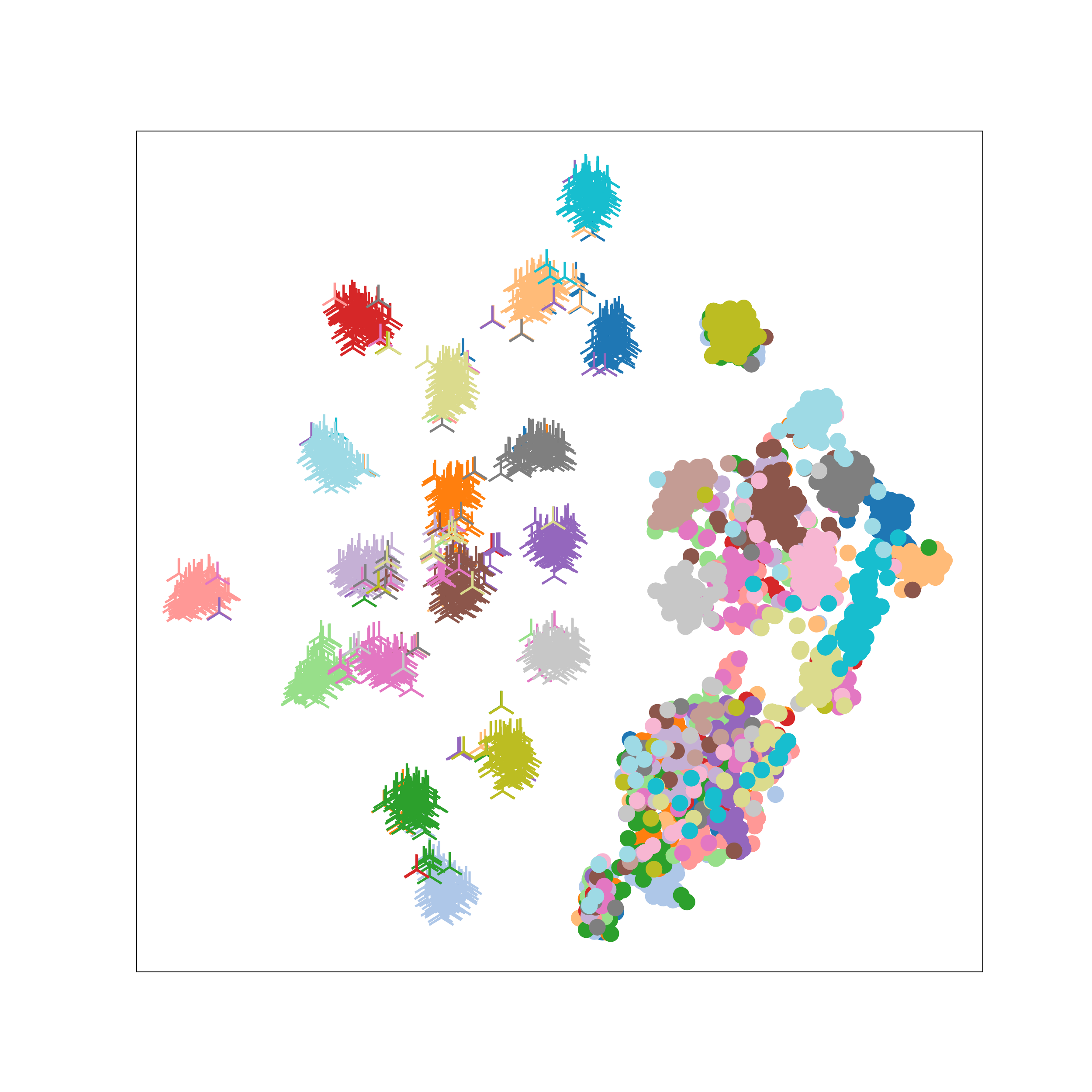}
        }
        \subfigure[Baseline+PL+P]{
            \includegraphics[width=0.25\textwidth]{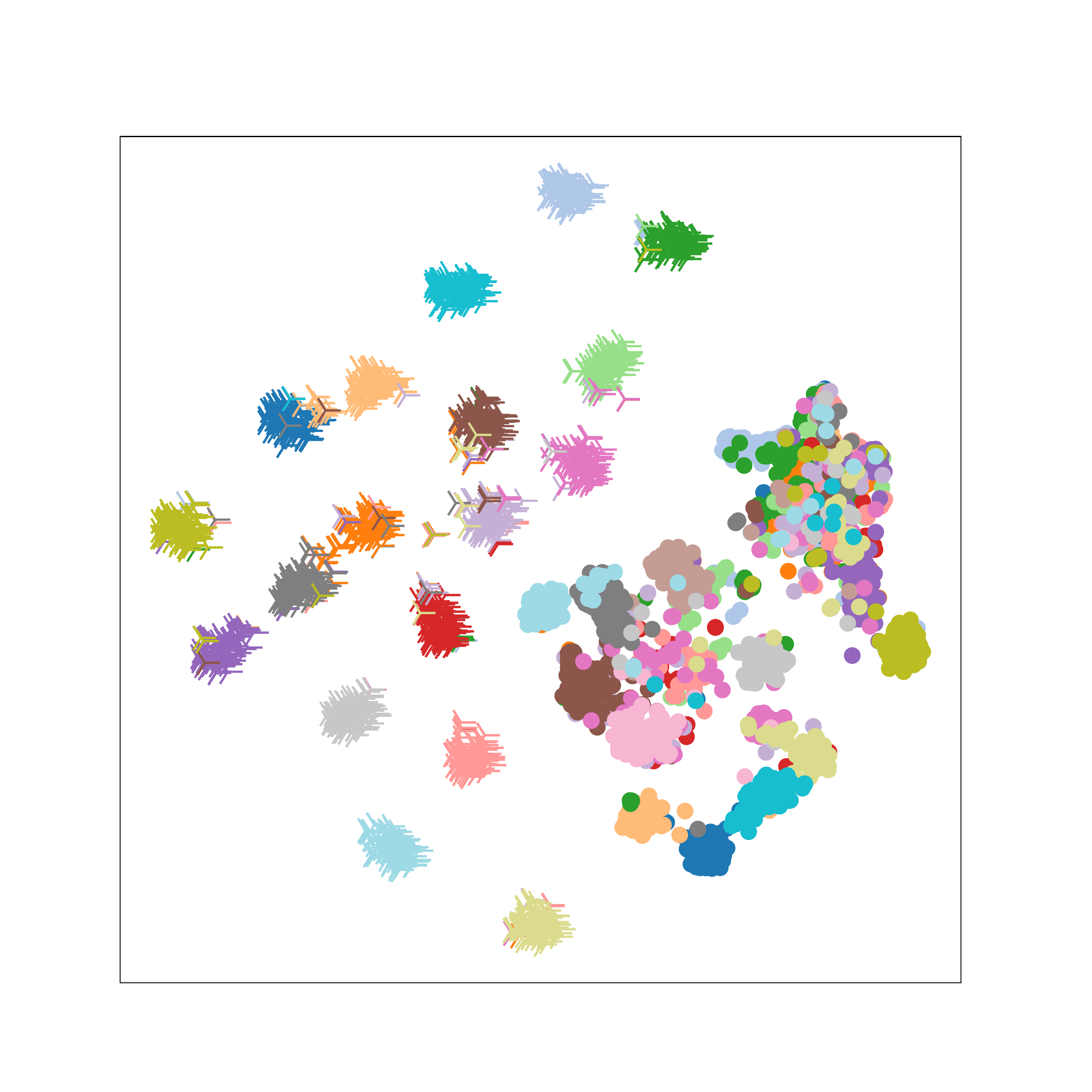}
        }
        \subfigure[Baseline+PL+C]{
            \includegraphics[width=0.25\textwidth]{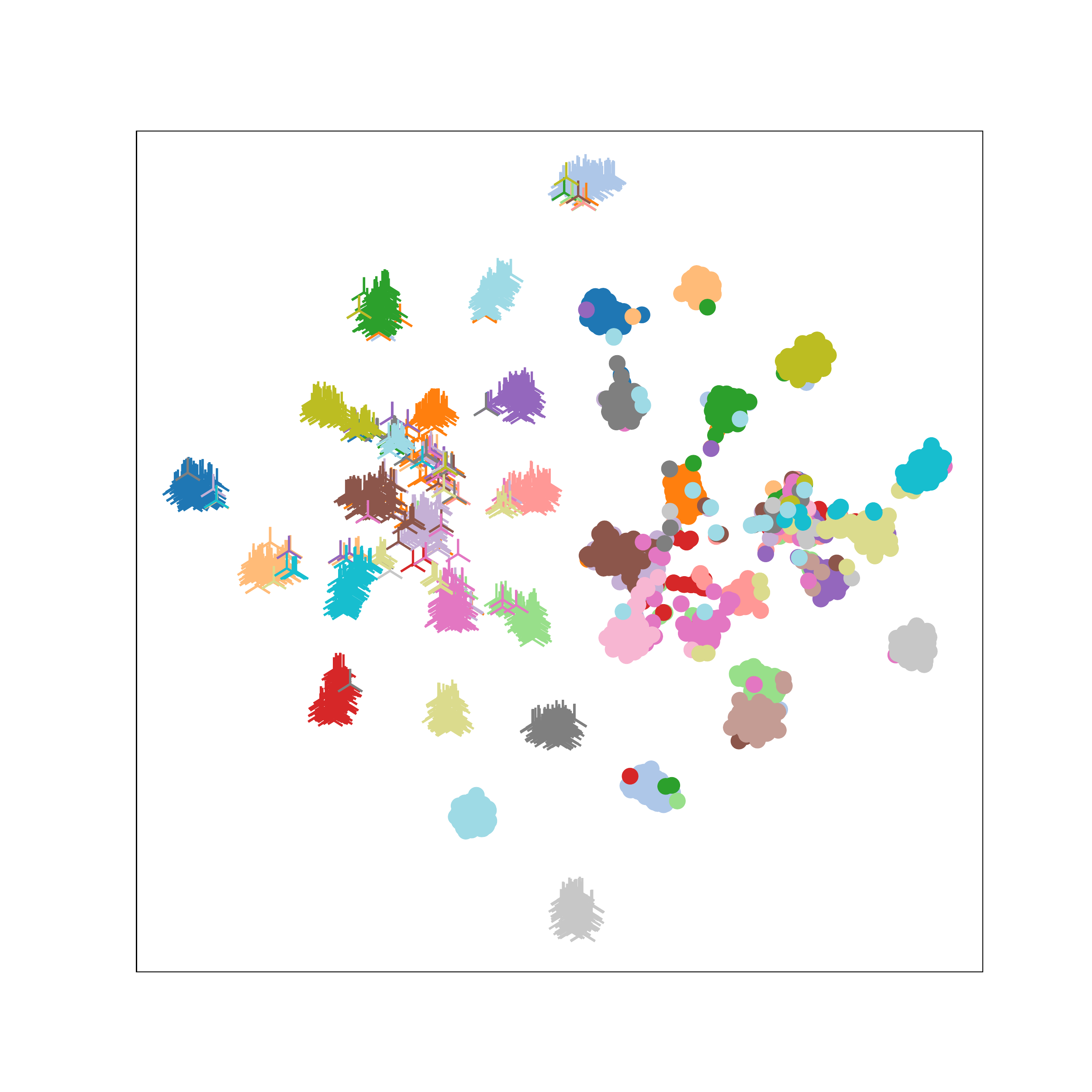}
        }
        \subfigure[Baseline+PL+C+D]{
            \includegraphics[width=0.25\textwidth]{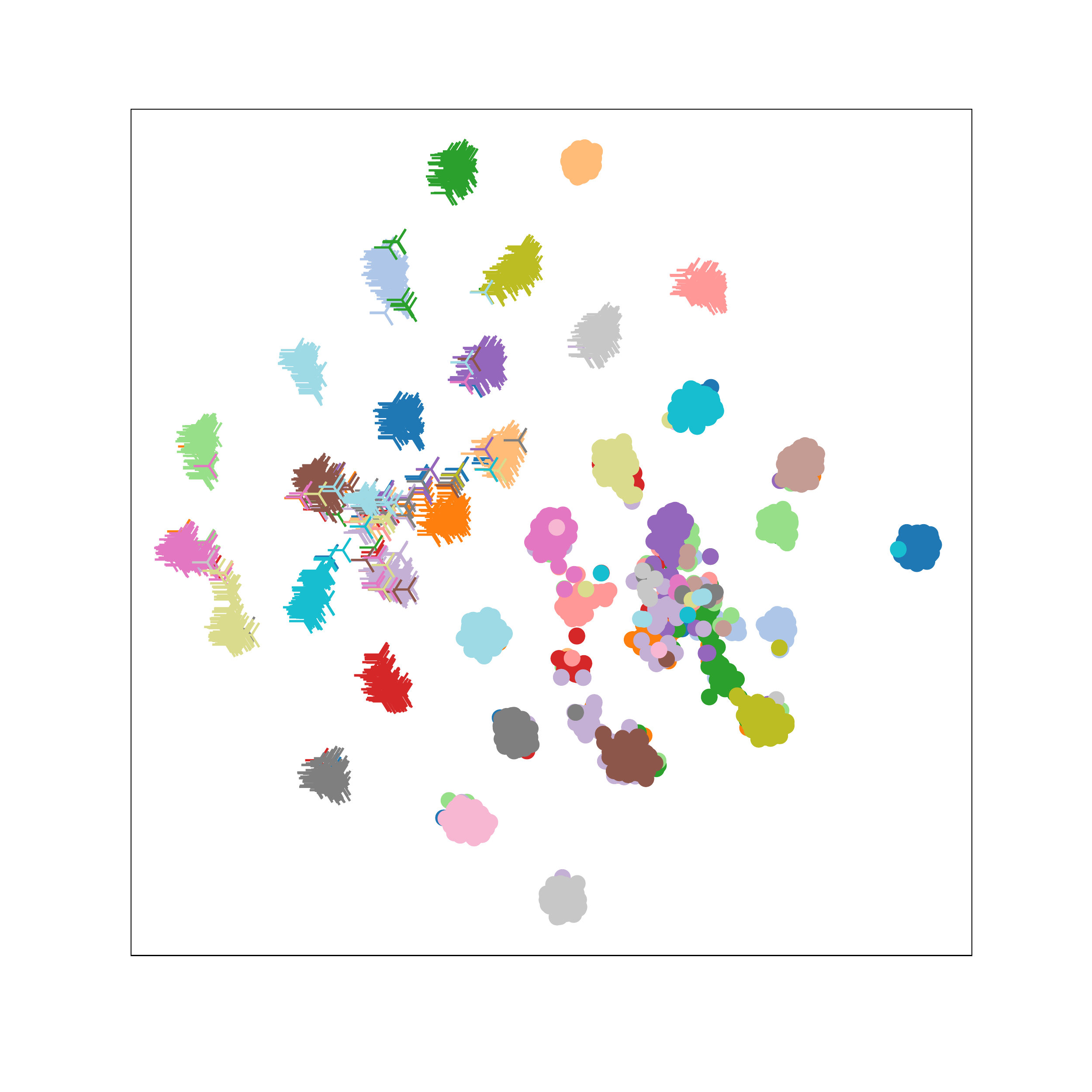}
        }
        \subfigure[Unbiased]{
            \includegraphics[width=0.25\textwidth]{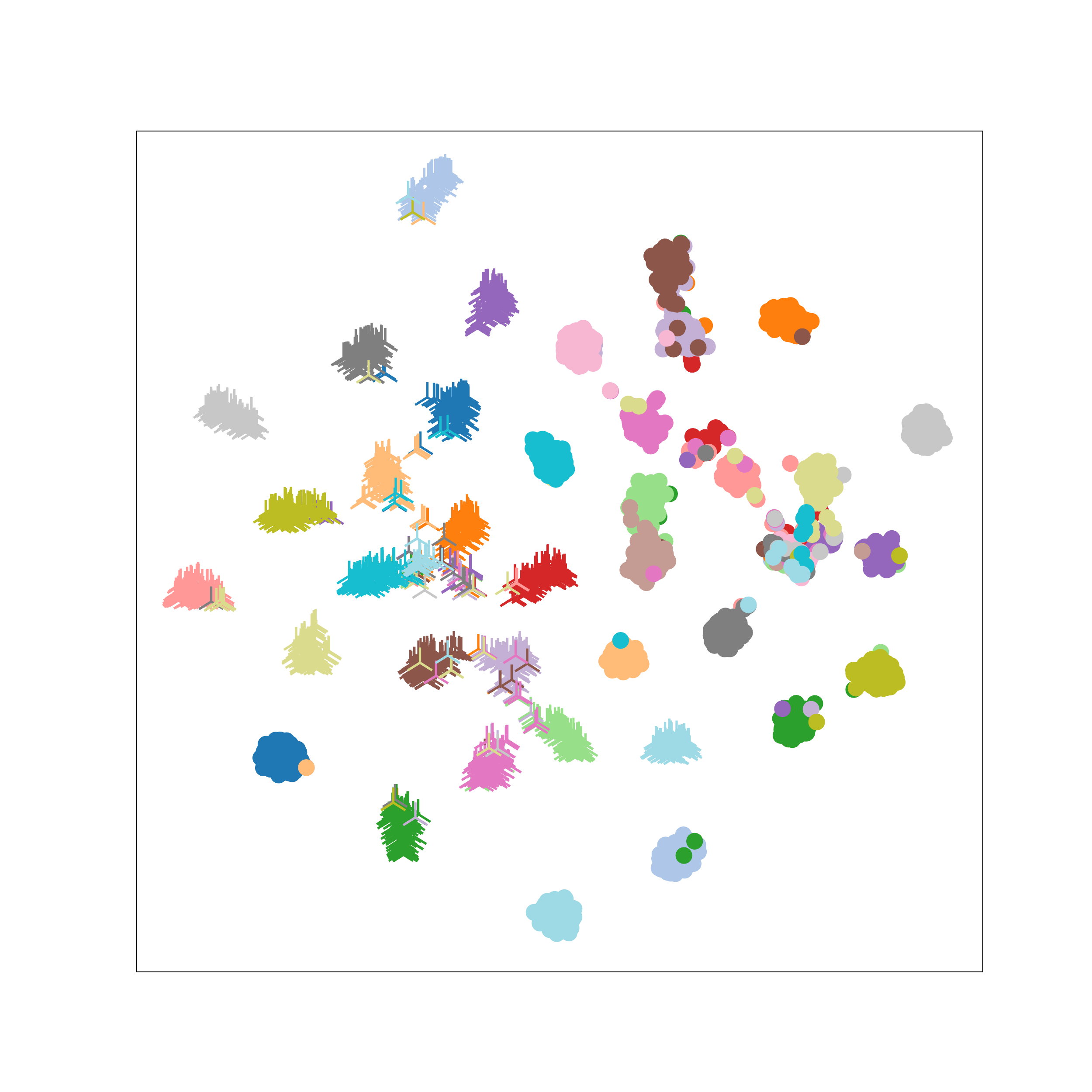}
        }
    }
    \caption{\textbf{T-SNE visualization before linear layer}: "PL, P, C, D" are the abbreviation of "Pseudo Label", "Position-Based Contrastive Learning", "Class-Based Contrastive Learning" and "Distance Aware Temperature". "Unbiased" represents using the ground truth of unseen classes for contrastive learning.  The "Dot" and "Triangle" markers in each figure denote the representations of the point clouds and the images from ImageNet, respectively. The distinct colors of markers indicate different categories.}
    \label{fig:TSNE_before}
\end{figure*}

\begin{figure*}[!t]
    \centering
    \resizebox{1\textwidth}{!}{
        \subfigure[Baseline]{
            \includegraphics[width=0.21\textwidth]{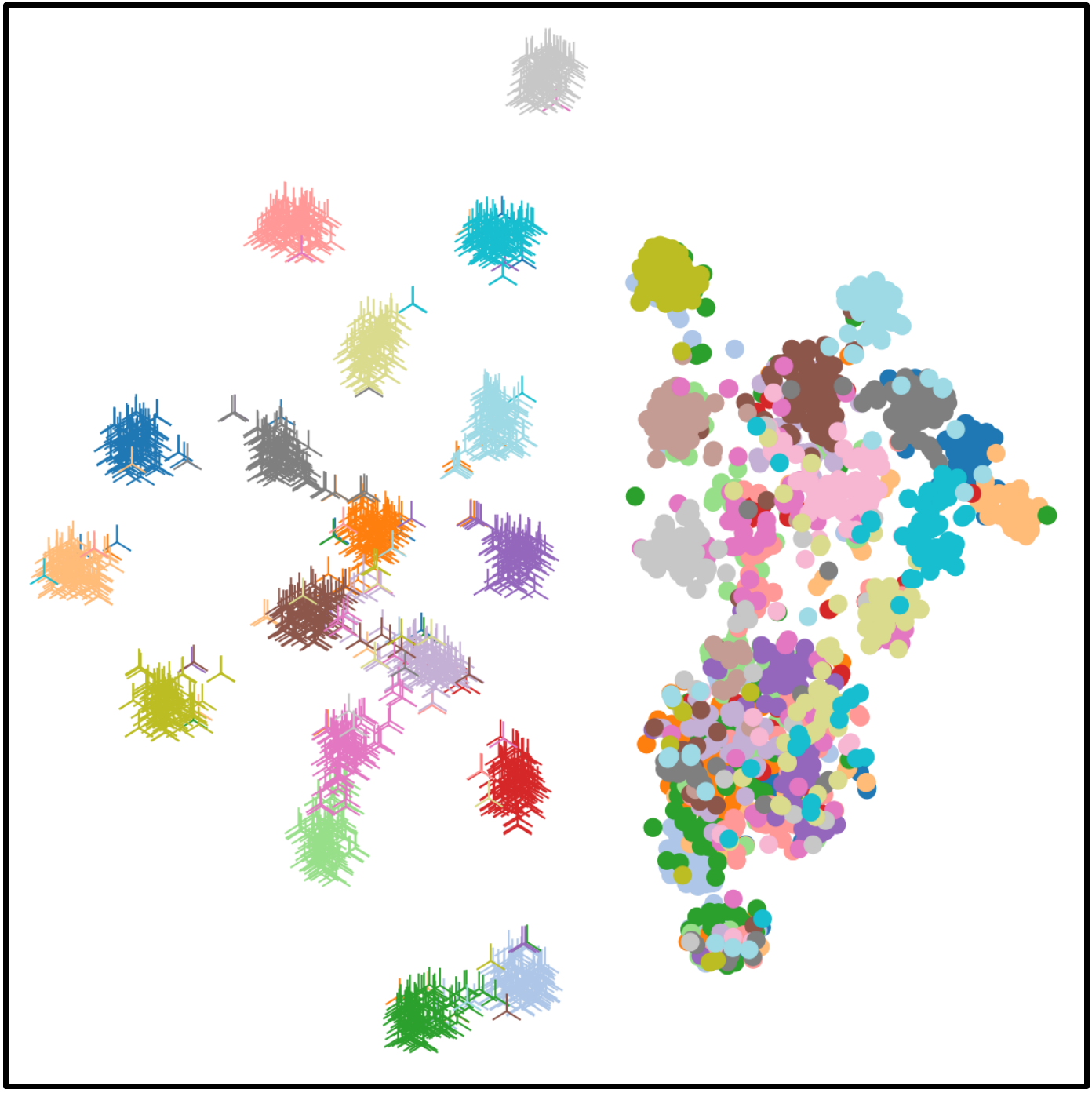}
        }
        \subfigure[Baseline+PL+P]{
            \includegraphics[width=0.21\textwidth]{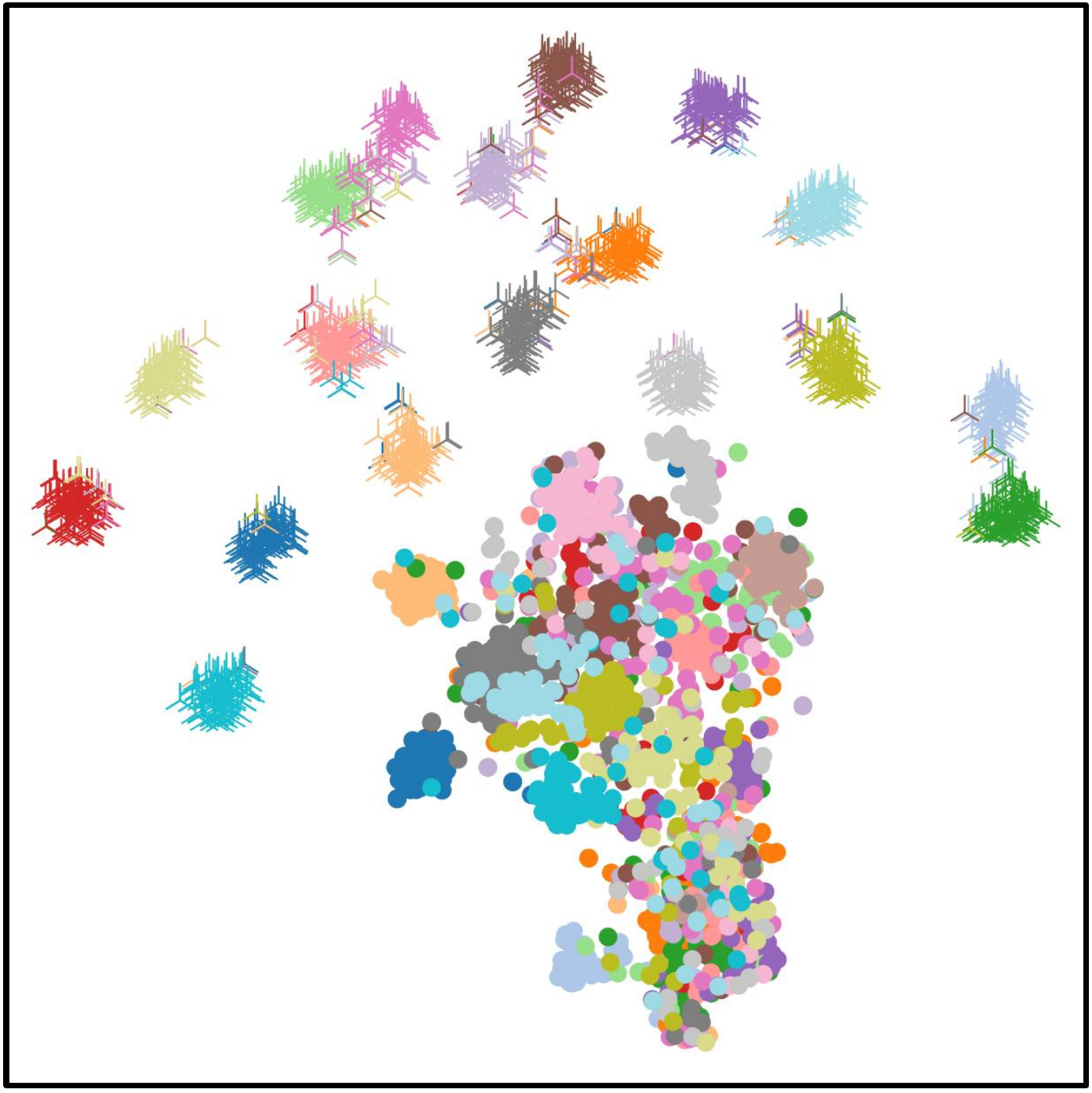}
        }
        \subfigure[Baseline+PL+C]{
            \includegraphics[width=0.21\textwidth]{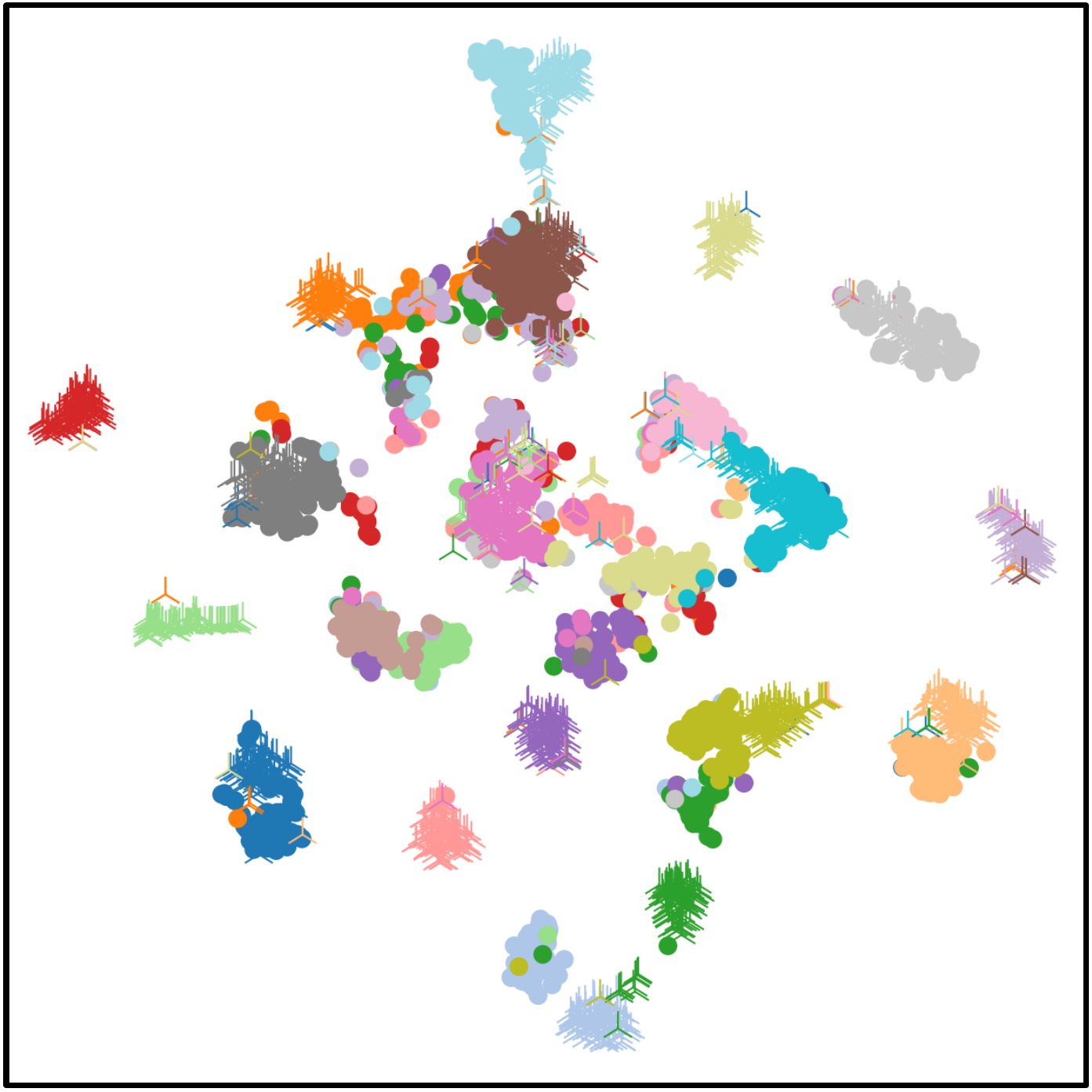}
        }
        \subfigure[Baseline+PL+C+D]{
            \includegraphics[width=0.21\textwidth]{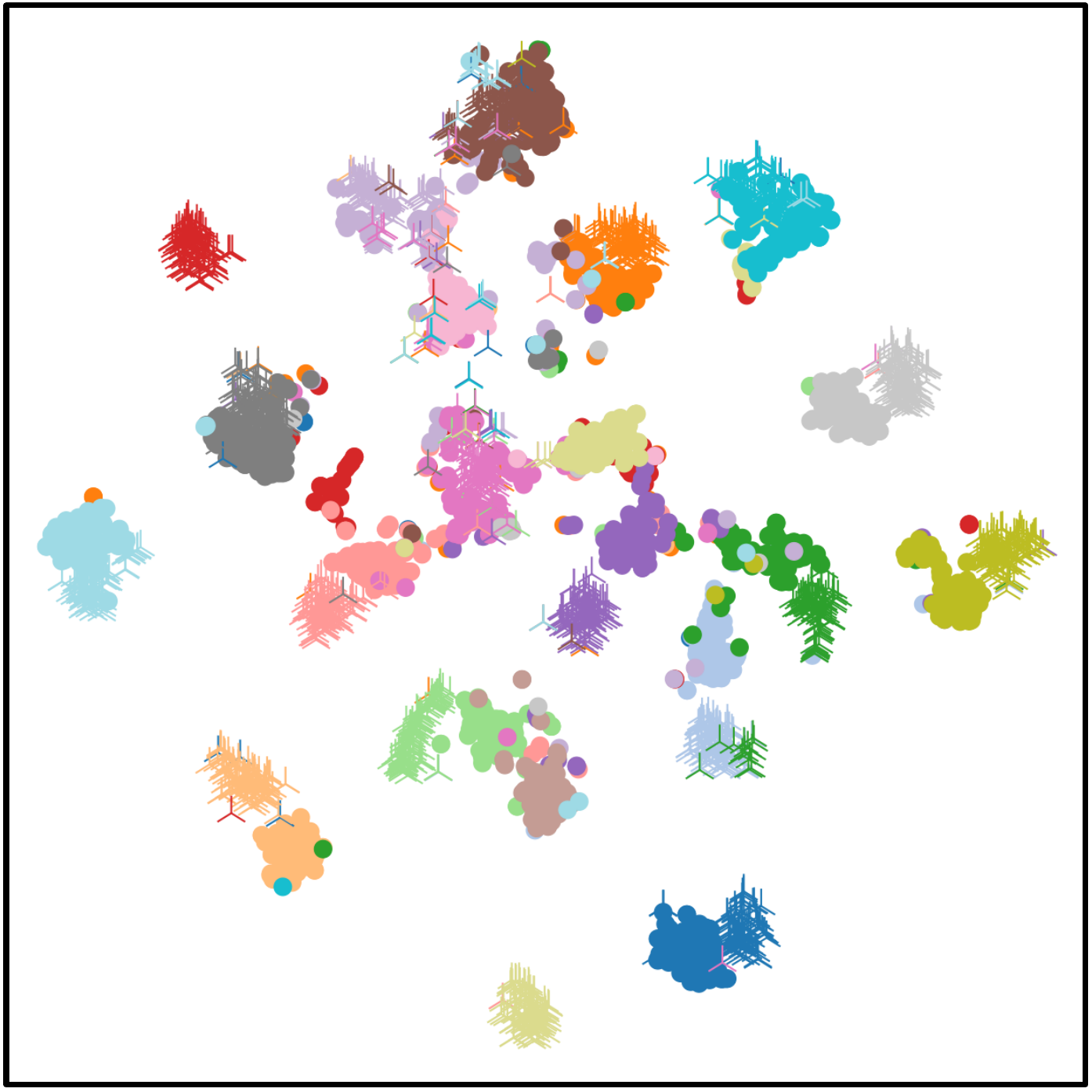}
        }
        \subfigure[Unbiased]{
            \includegraphics[width=0.21\textwidth]{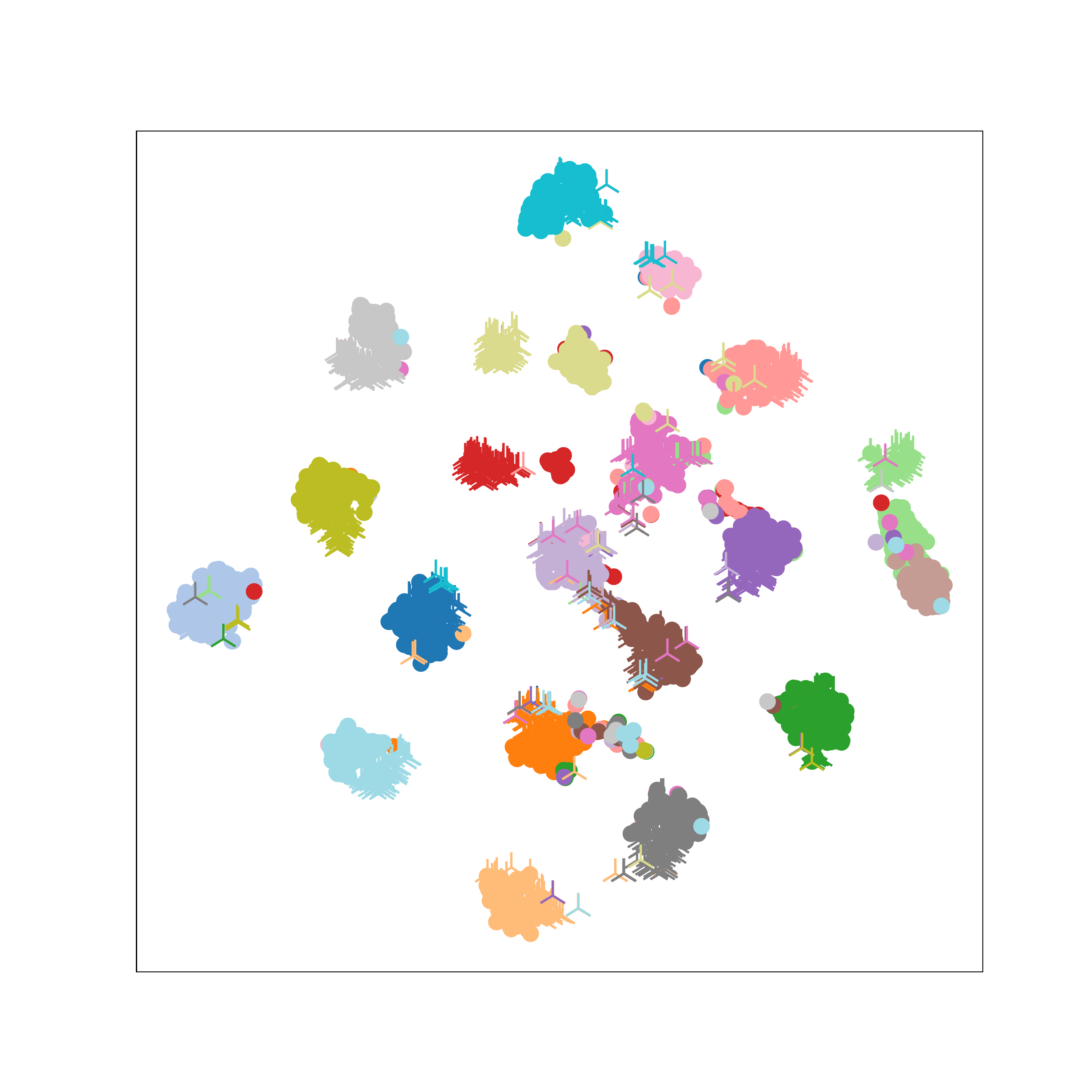}
        }
    }
    \caption{\textbf{T-SNE visualization after linear layer}: "PL, P, C, D" are the abbreviation of "Pseudo Label", "Position-Based Contrastive Learning", "Class-Based Contrastive Learning" and "Distance Aware Temperature". "Unbiased" represents using the ground truth of unseen classes for contrastive learning. The "Dot" and "Triangle" markers in each figure denote the representations of the point-clouds and the images from ImageNet, respectively. The distinct colors of markers indicate different categories.}
    \label{fig:TSNE_after}
\end{figure*}


\paragraph{Analysis of Learned Representations} Inspired by \cite{chuang2020debiased}, we investigate the learned representation via comparison among T-SNEs. Specifically, we compare OV-3DETIC with four different settings, they are baseline setting, baseline and pseudo label with position-based contrastive learning, baseline and pseudo label with class-based contrastive learning, and unbiased contrastive learning with ground truth. Please note that the unbiased contrastive learning is the upper bound of OV-3DETIC. Besides, we visualize the feature both before and after the linear layer of contrastive loss. The results are shown in Figure. \ref{fig:TSNE_before} and Figure. \ref{fig:TSNE_after}. The "Dot" and "Triangle" markers denote the feature of point-cloud and images from ImageNet, respectively. Figure \ref{fig:TSNE_before} (a) can be mainly divided into three parts: ImageNet features (left), seen classes features of point-clouds (right-top) and unseen classes features of point-clouds (right-down). The comparison among these three parts indicates that, in the baseline setting, ImageNet features are clearly clustered, and that of seen classes could barely distinguish from each other, while those of unseen classes are almost indistinguishable. Moreover, compare the first four sub-figures of Figure. \ref{fig:TSNE_before}, we find that the "Pseudo label", "Class-based contrastive learning" and "Distance-aware temperature" help the clustering of the representations of unseen classes progressively. 

Above observations can be also found in the feature distribution after the linear layer, we can evaluate the gain of de-biased contrastive learning by observing the ImageNet and point cloud feature distributions. Specifically, if these two feature distributions are close to each other, then the contrastive strategy works. As shown in the Figure. \ref{fig:TSNE_after} (e), in the unbiased setting, the two feature distributions coincide with each other, and in the biased setting (Figure. \ref{fig:TSNE_after} (b)), there is a significant difference between these two distributions. our de-biased setting (Figure. \ref{fig:TSNE_after} (d)) performs better than the biased setting, while there are still some hard examples that could be improved.

\begin{figure*}[!t]
    \centering
    \subfigure[Performance vs. Data Ratio.]{
        \label{fig:Performance vs data amount}
        \begin{minipage}[t]{0.45\linewidth}
            \centering
            \begin{tikzpicture}[scale=0.5]
                \centering
            	\begin{axis}[
            	    axis lines*=left,
            	    ymajorgrids = true,
            	    ylabel=$mAP_{25}$ (\%),
            	    xlabel=Data Ratio (\%),
                    xmin=0,ymin=0,ymax=50,
                    xtick={0,10,30,50,70,100},
                    ytick={0,10,20,30,40,50},
                    grid=major,
            	]
                    \addplot[draw=ForestGreen,mark=x] 
                    coordinates {
                        (1,0.08)
                        (10,12.24)
                        (30,17.38)
                        (70,28.87)
                        (100,31.84)
                    };
                    \label{ScanNet-Point-cloud}
                    \addplot[draw=BurntOrange,mark=x]
                    coordinates {
                        (1,0.4)
                        (10,4.5)
                        (30,12.4)
                        (70,38.1)
                        (100,45.7)
                    };
                    \label{ScanNet-Image}
                \end{axis}
            	\begin{axis}[
            	    axis y line*=right,
            	    axis x line=none,
            	    ylabel=$AR_{25}$ (\%),
                    xmin=0,ymin=0,ymax=80,
                    ytick={0,10,20,30,40,50,60,70,80},
                    legend style={at={(0.65,0.40)},anchor=north,legend columns=1},
            	]
            	    \addlegendimage{/pgfplots/refstyle=ScanNet-Point-cloud}\addlegendentry{Point-cloud $mAP_{25}$}
            	    \addlegendimage{/pgfplots/refstyle=ScanNet-Image}\addlegendentry{Image $mAP_{25}$}
                    \addlegendentry{Point-cloud $AR_{25}$}
                    \addplot[draw=ForestGreen,no marks, dashed] 
                    coordinates {
                        (1,28.93)
                        (10,51.65)
                        (30,56.38)
                        (70,59.98)
                        (100,61.84)
                    };
                    \addlegendentry{Image $AR_{25}$}
                    \addplot[draw=BurntOrange,no marks, dashed]
                    coordinates {
                        (1,7.1)
                        (10,38.7)
                        (30,54.4)
                        (70,69.1)
                        (100,74.5)
                    };
                \end{axis}
            \end{tikzpicture}
        \end{minipage}%
    }%
    \subfigure[Performance vs. Iteration.]{
        \label{fig:Performance vs iteration}
        \begin{minipage}[t]{0.45\linewidth}
            \centering
            \begin{tikzpicture}[scale=0.5]
                \centering
            	\begin{axis}[
            	    axis lines*=left,
            	    ymajorgrids = true,
                    xlabel=Iteration,
            	    ylabel=$mAP_{25}$ (\%),
                    xmin=0,ymin=0,ymax=14,
                    xtick={0,1,2,3,4},
                    ytick={0,2,4,6,8,10,12,14},
                    grid=major,
            	]
                    \addplot[draw=ProcessBlue,mark=x] 
                    coordinates {
                        (0,1.157)
                        (1,6.051)
                        (2,10.075)
                        (3,11.846)
                        (4,12.55)
                    };
                    \label{plot_one}
                \end{axis}
            	\begin{axis}[
            	    axis y line*=right,
            	    axis x line=none,
                    xlabel=Iteration,
            	    ylabel=$AR_{25}$ (\%),
                    xmin=0,ymin=0,ymax=45,
                    ytick={0,5,10,15,20,25,30,35,40,45},
                    legend style={at={(0.75,0.25)},anchor=north,legend columns=1},
            	]
            	    \addlegendimage{/pgfplots/refstyle=plot_one}\addlegendentry{$mAP_{25}$}
                    \addlegendentry{$AR_{25}$}
                    \addplot[draw=BrickRed,mark=x] 
                    coordinates {
                        (0,27.267)
                        (1,33.52)
                        (2,36.391)
                        (3,40.337)
                        (4,38.353)
                    };
                    
                \end{axis}
            \end{tikzpicture}
        \end{minipage}%
    }%
    \centering
\label{}
\caption{(a). illustrates relation between $mAP_{25}$, $AR_{25}$ and the data regime of both 2D and 3D detection, the green, orange, solid and dashed lines represent point cloud, image, $mAP_{25}$ and $AR_{25}$, respectively. (b). illustrates the relation between $mAP_{25}$, $AR_{25}$ and pseudo label iteration, the blue and red lines represents $mAP_{25}$ and $AR_{25}$, respectively.}
\end{figure*}
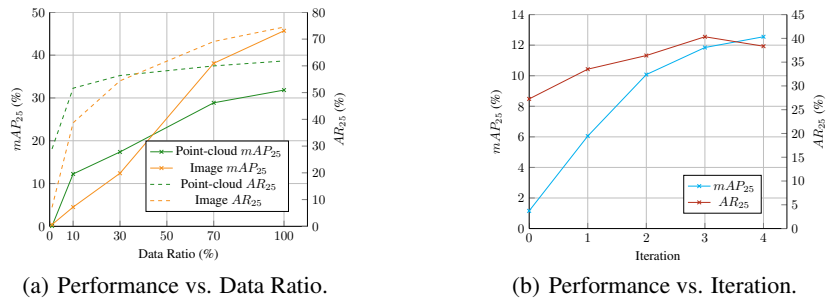

\paragraph{Analysis of Performance on Different Training Data Ratio}
Figure. \ref{fig:Performance vs data amount} illustrate the relation between $mAP_{25}$, $AR_{25}$ and the different training data ratio. The left $y$ axis denotes $mAP_{25}$, the right $y$ axis denotes $AR_{25}$, and the $x$ axis represents the data ratio used during training. We draw both $mAP_{25}$, $AR_{25}$ results of the point cloud and the paired image branch. On one hand, compared with $mAP_{25}$, $AR_{25}$ converges with comparatively less training data, which occurs in both the image and the point cloud branches. $AR_{25}$ partially reflects localization ability, which means we can train a generalizable bounding box detector with few annotations. On the other hand, compared with $AR_{25}$ of the image branch, the $AR_{25}$ of the point cloud branch converges with less training data, which means the localization ability of the 3D detector is better than the 2D detector, that is why we use the point-cloud detector to generate bounding box pseudo labels. Moreover, the converged $mAP_{25}$ of the image branch is better than the point cloud branch. This may be because texture and detail account for a lot of classification, which is the weakness of the point cloud detector, however.

\paragraph{Analysis on Pseudo Label Effects}
Figure. \ref{fig:Performance vs iteration} illustrates the relation between $mAP_{25}$, $AR_{25}$ and pseudo label iteration. During the second phase, we iteratively update the pseudo label every 50 epochs. The results indicate that the more iterations, the better the pseudo labels should be, which leads to better performance.

\subsection{Qualitative Results}
\begin{figure*}[htbp]
	\centering
	\vspace{-3mm}
	\setcounter{subfigure}{0}

	\subfigure{
        \rotatebox{90}{\scriptsize{~~~~~~~~~~~~Baseline}}
		\begin{minipage}[t]{0.2\linewidth}
			\centering
			\raisebox{0.0\height}{\includegraphics[width=1\linewidth]{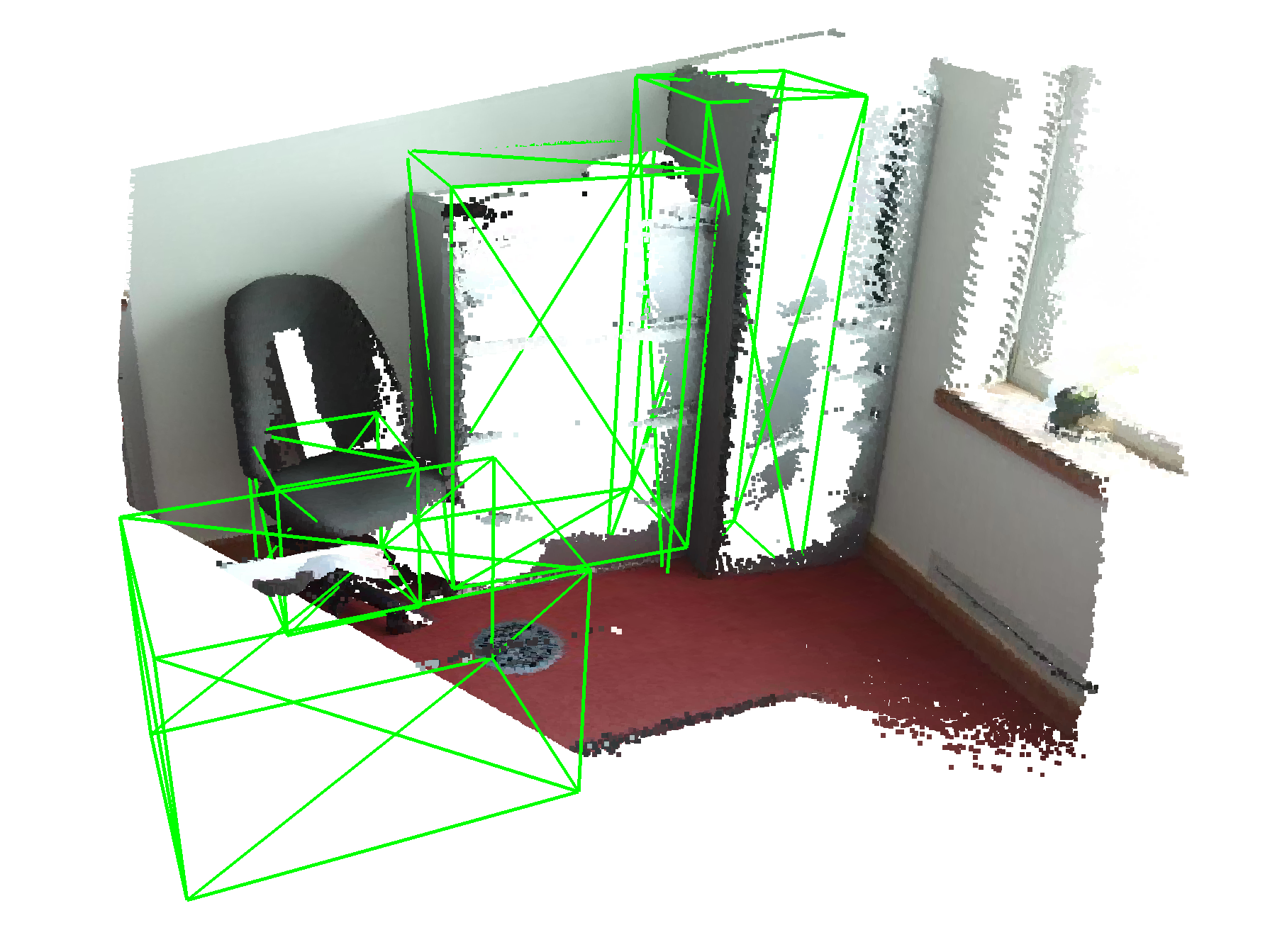}}
		\end{minipage}
	}
	\subfigure{
		\begin{minipage}[t]{0.2\linewidth}
			\centering
			\raisebox{0.0\height}{\includegraphics[width=1\linewidth]{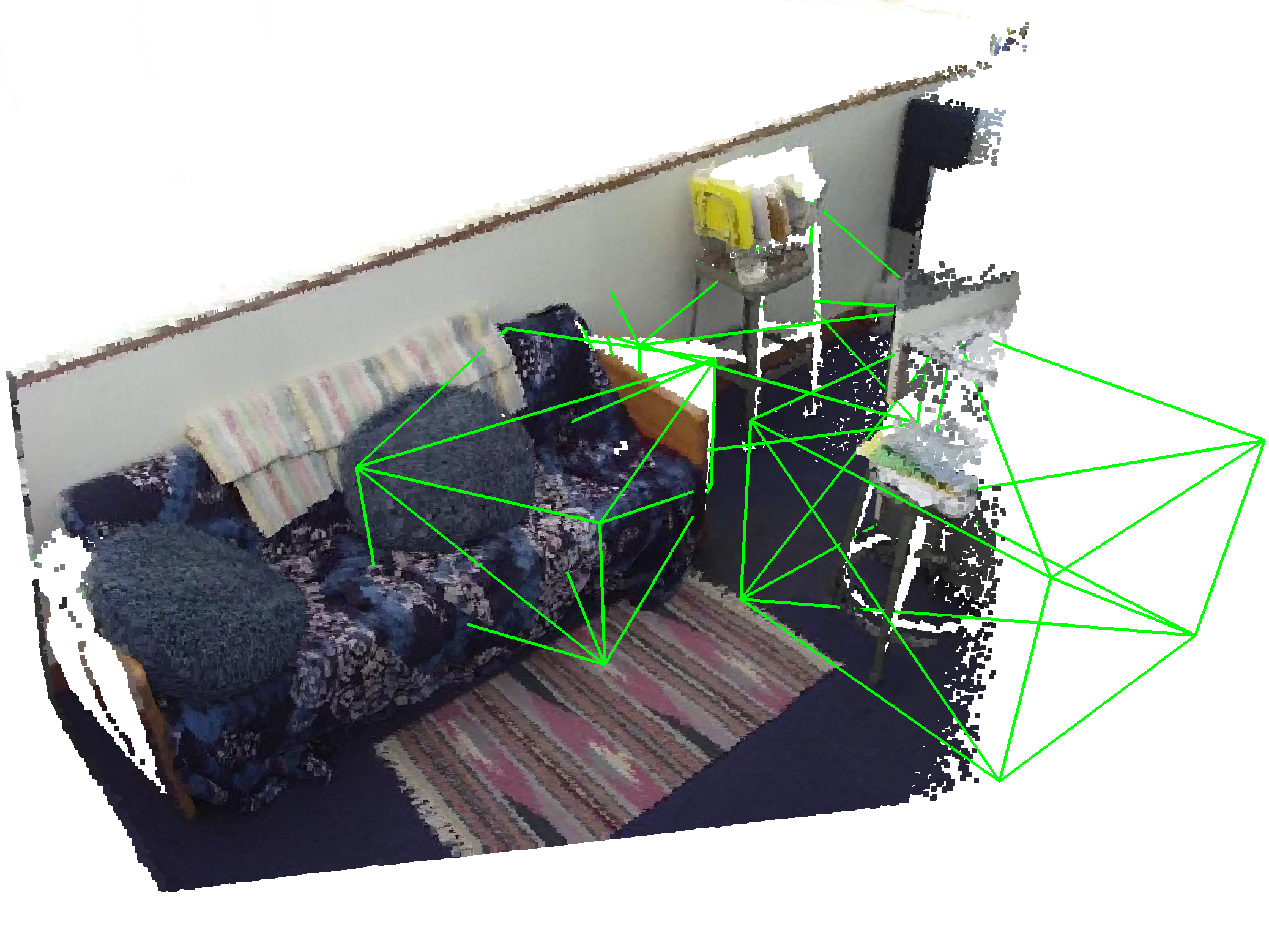}}
		\end{minipage}

	}
	\subfigure{
		\begin{minipage}[t]{0.2\linewidth}
			\centering
			\raisebox{0.0\height}{\includegraphics[width=1\linewidth]{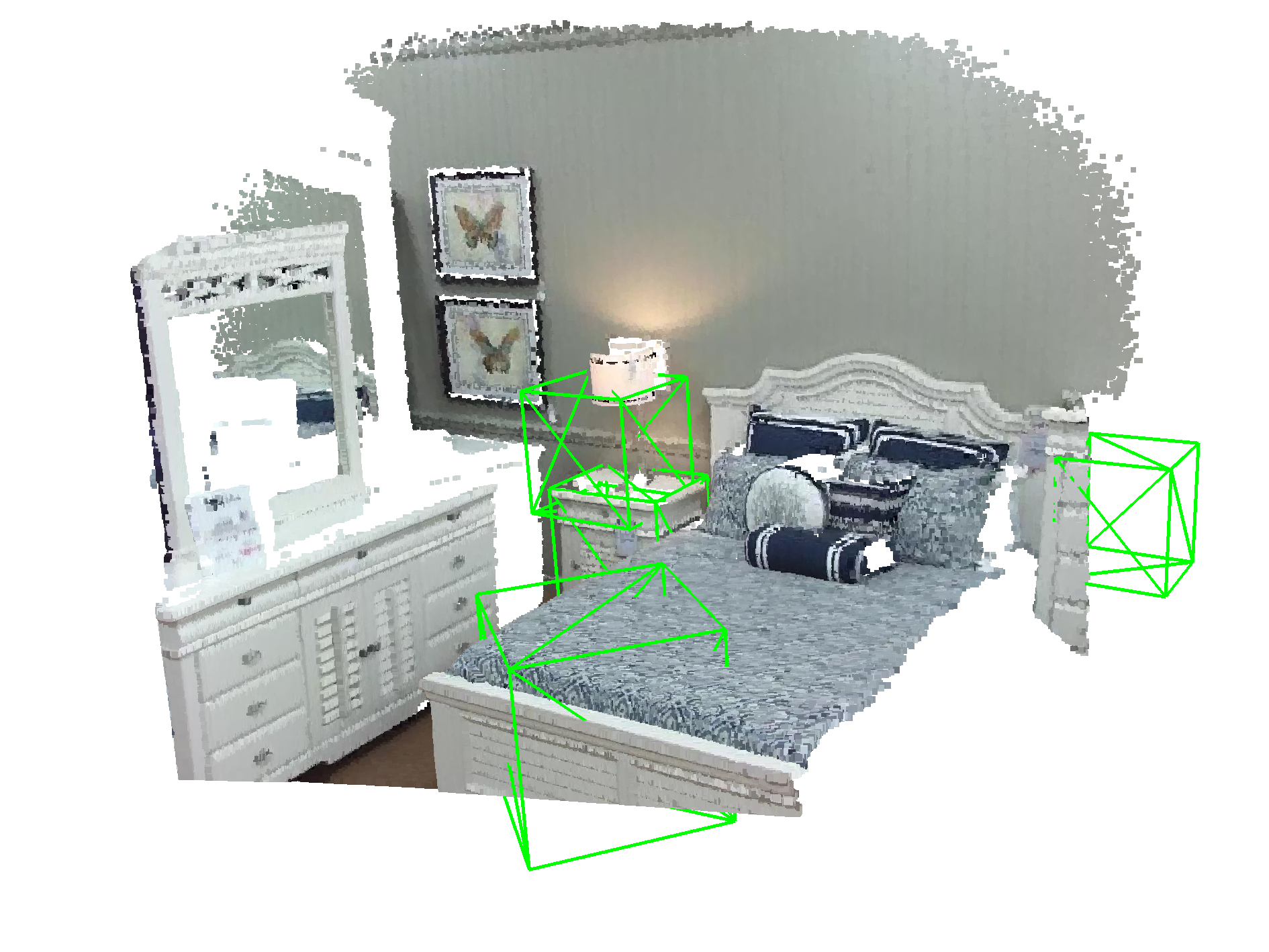}}
		\end{minipage}
	}
	\subfigure{
		\begin{minipage}[t]{0.2\linewidth}
			\centering
			\raisebox{0.0\height}{\includegraphics[width=1\linewidth]{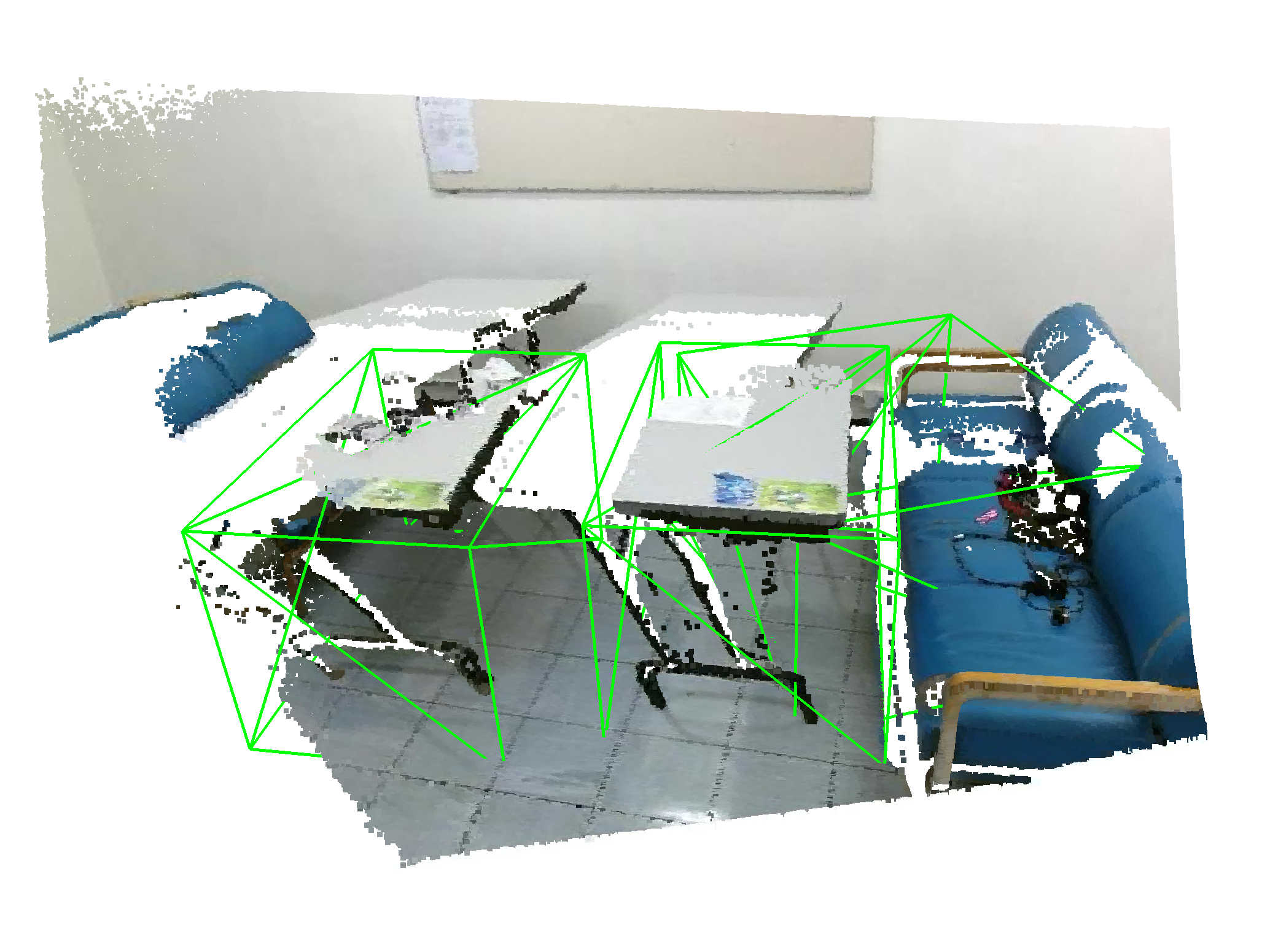}}
		\end{minipage}
	}
	
	\vspace{-5mm}
	\subfigure{
        \rotatebox{90}{\scriptsize{~~~~~~~~~~~~~OV-3DETIC}}
		\begin{minipage}[t]{0.2\linewidth}
			\centering
			\raisebox{0.0\height}{\includegraphics[width=1\linewidth]{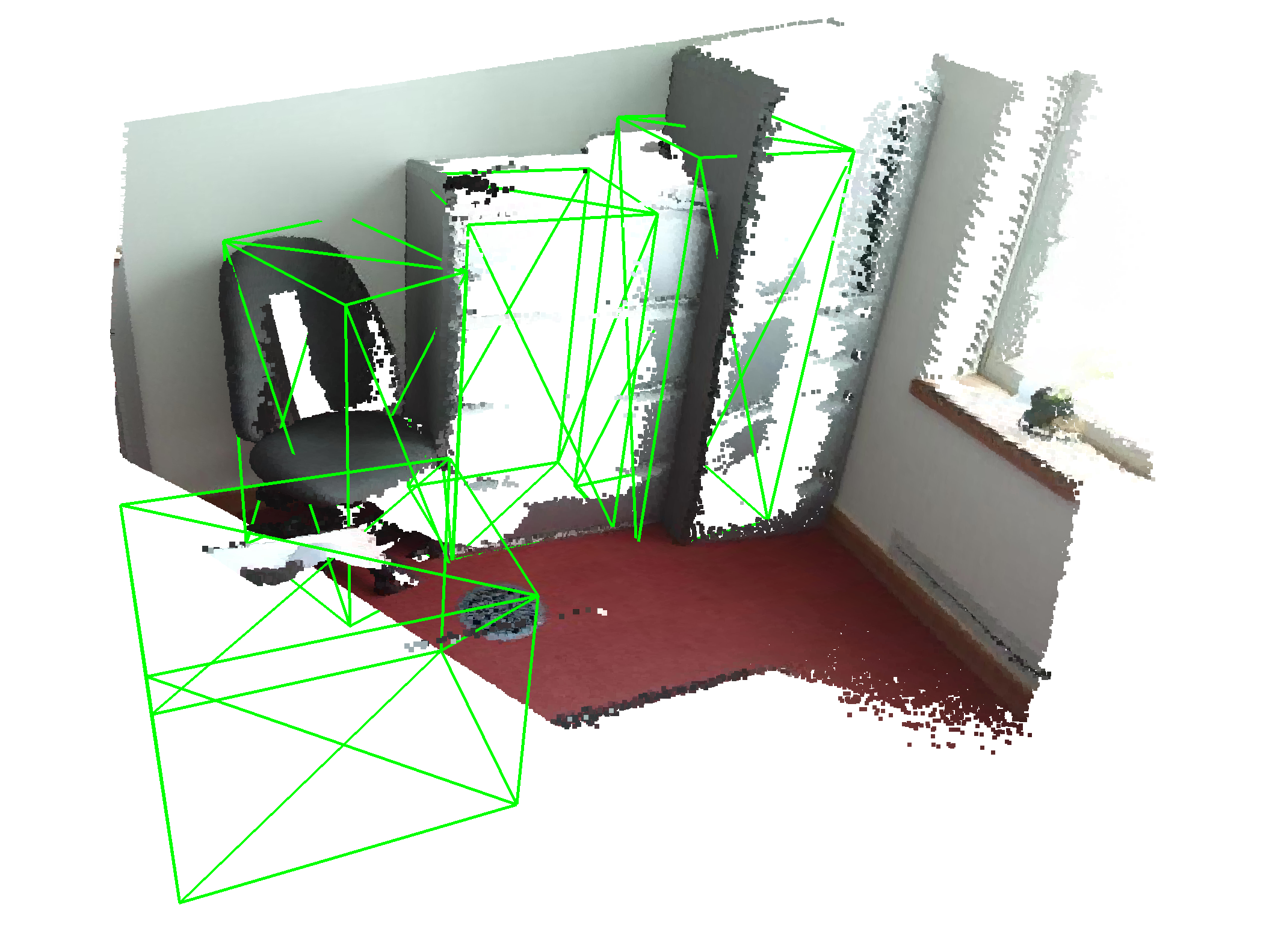}}
		\end{minipage}
	}
	\subfigure{
		\begin{minipage}[t]{0.2\linewidth}
			\centering
			\raisebox{0.0\height}{\includegraphics[width=1\linewidth]{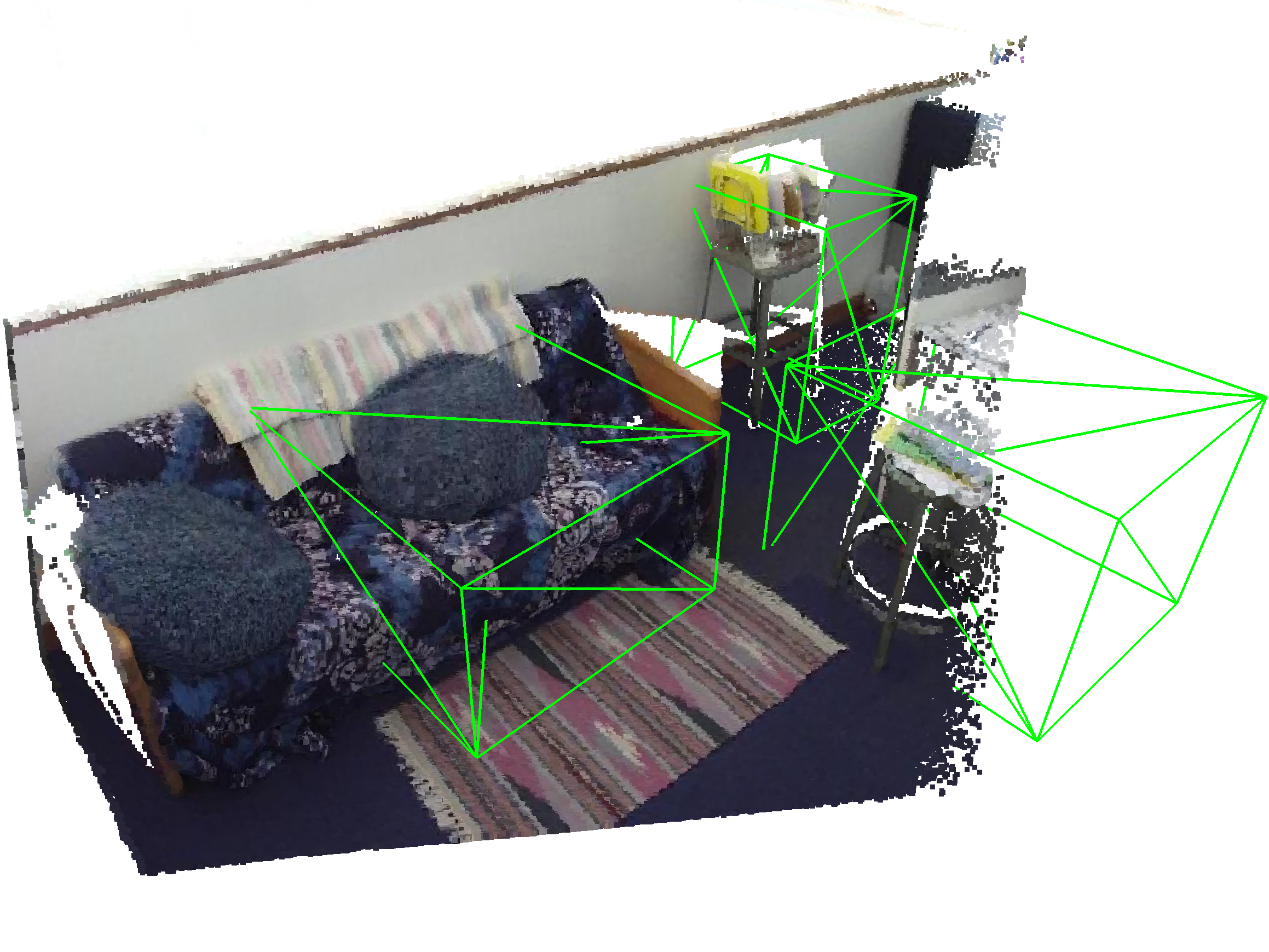}}
		\end{minipage}
	}
	\subfigure{
		\begin{minipage}[t]{0.2\linewidth}
			\centering
			\raisebox{0.0\height}{\includegraphics[width=1\linewidth]{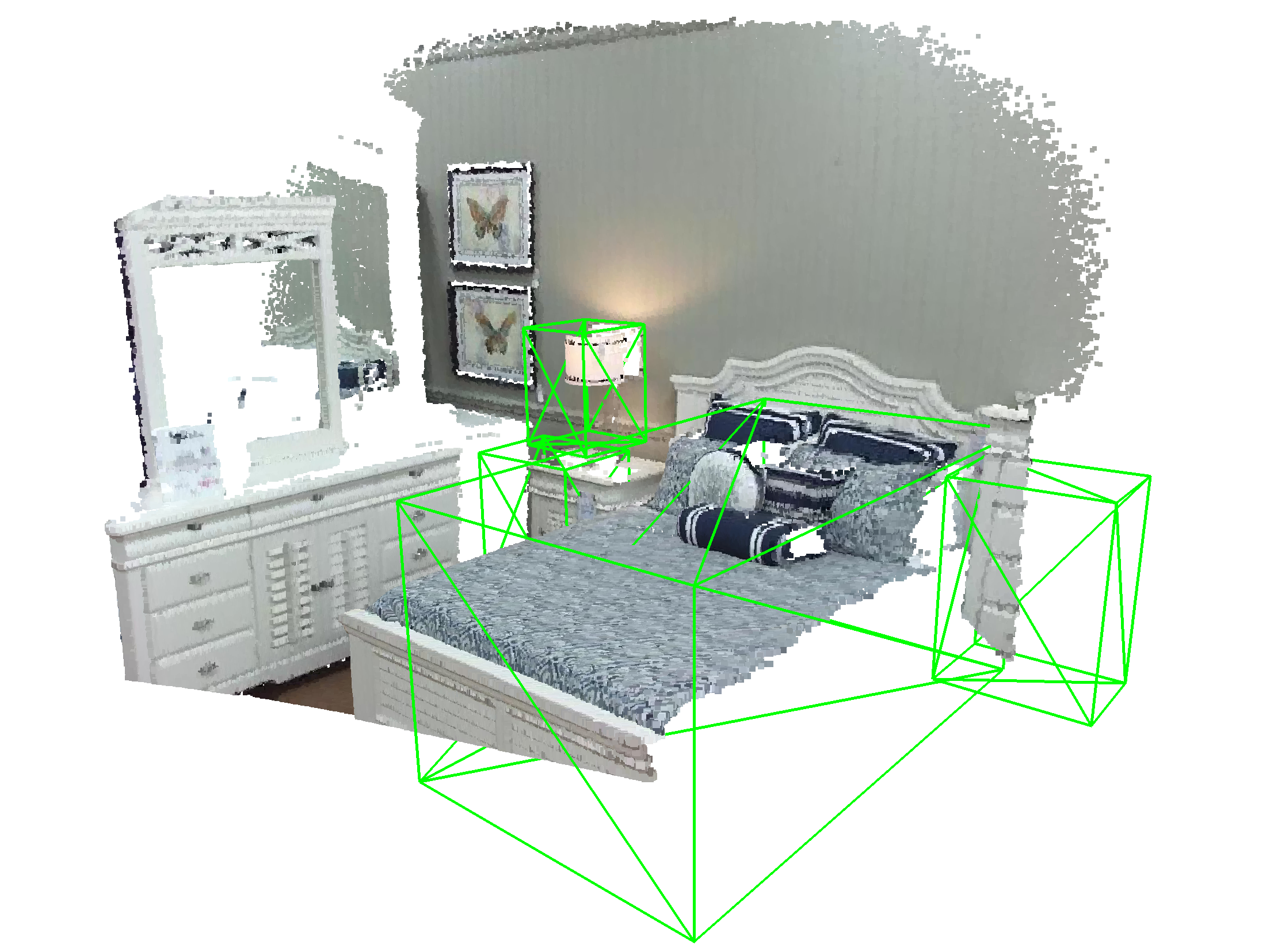}}
		\end{minipage}
	}
	\subfigure{
		\begin{minipage}[t]{0.2\linewidth}
			\centering
			\raisebox{0.0\height}{\includegraphics[width=1\linewidth]{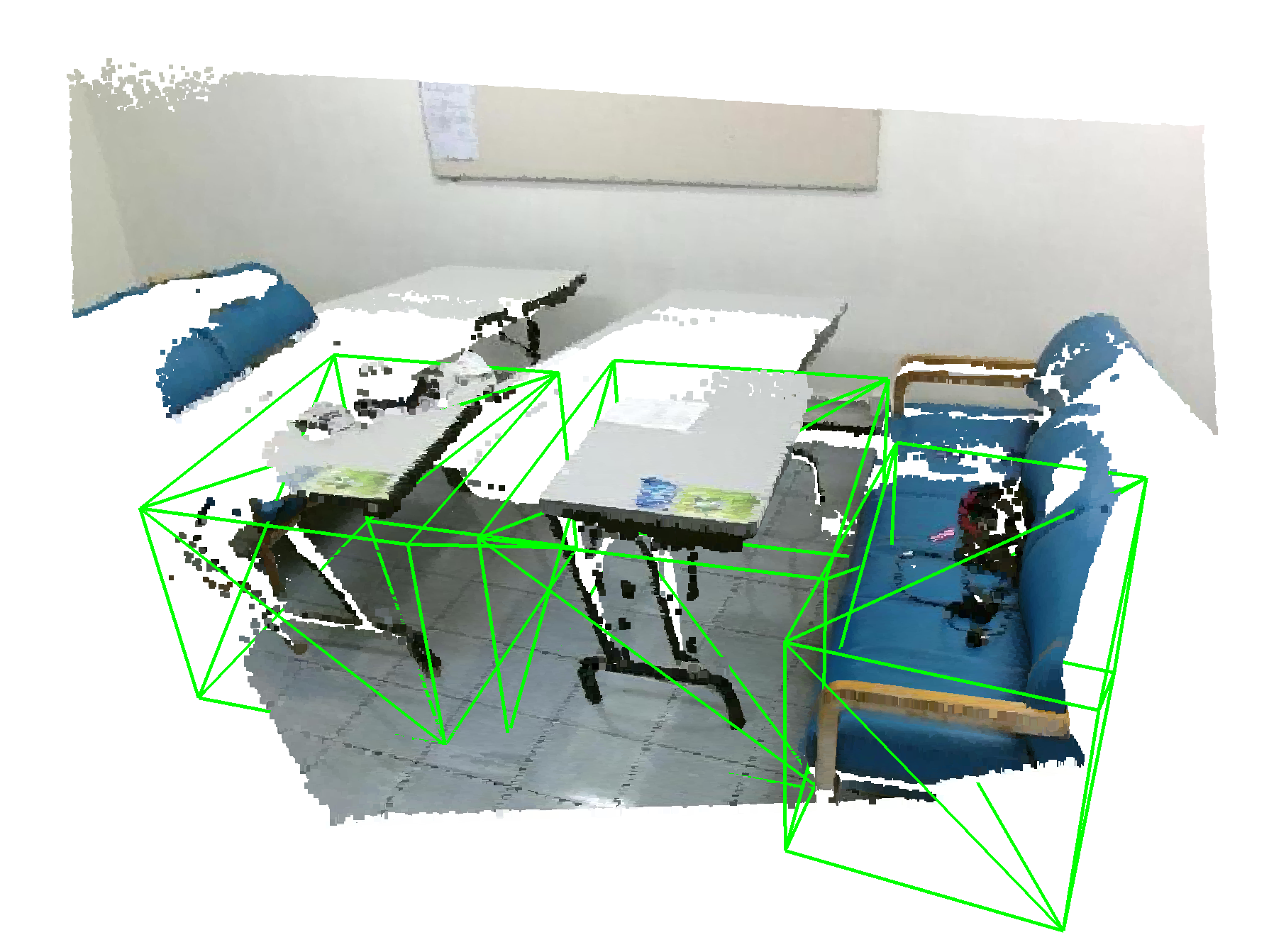}}
		\end{minipage}
	}
	
	\vspace{-5mm}
	
	\setcounter{subfigure}{0}

	\subfigure[Scene 1]{
        \rotatebox{90}{\scriptsize{~~~~~~~~~~~Ground Truth}}
		\begin{minipage}[t]{0.2\linewidth}
			\centering
			\raisebox{0.0\height}{\includegraphics[width=1\linewidth]{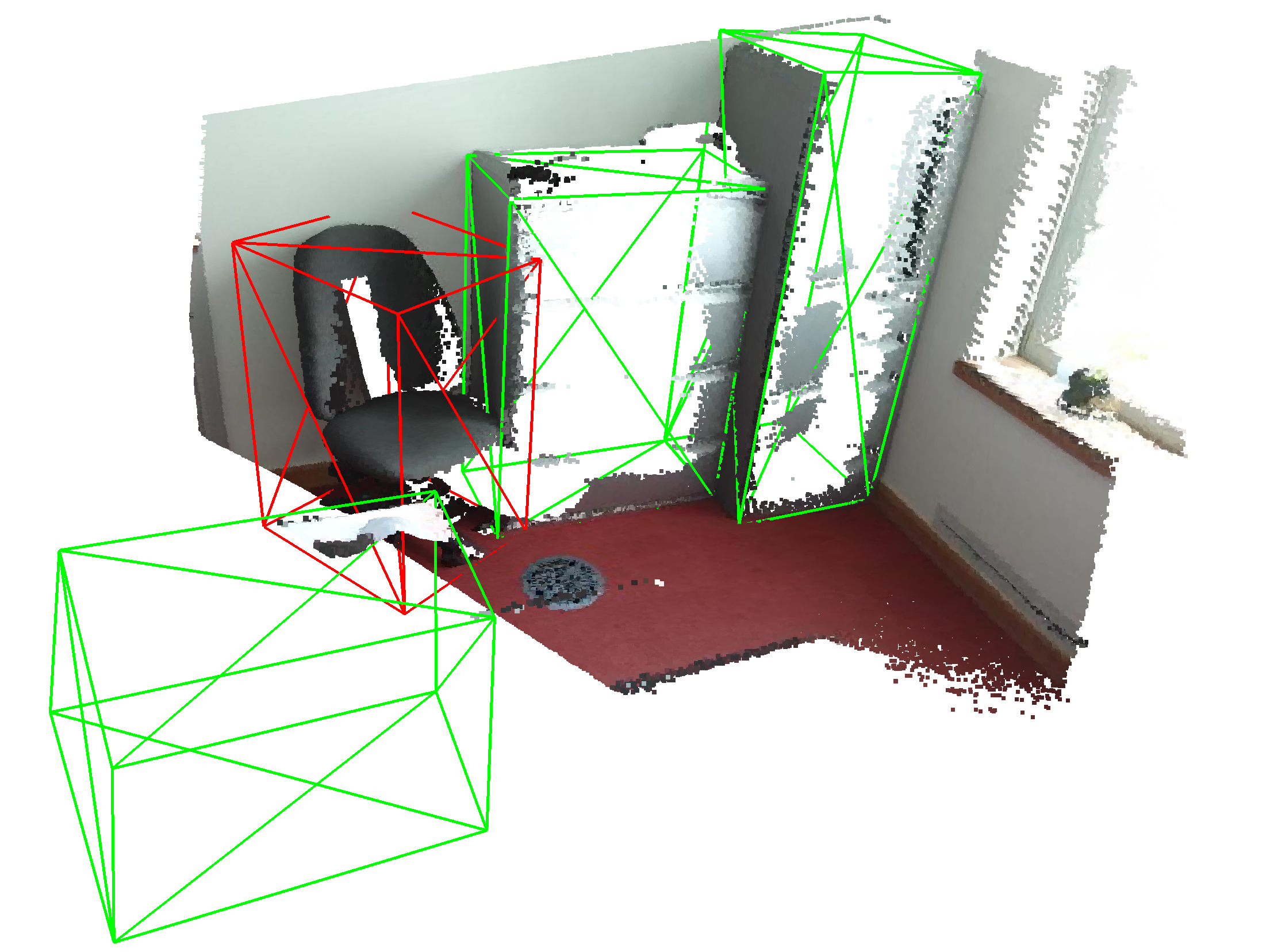}}
		\end{minipage}
	}
	\subfigure[Scene 2]{
		\begin{minipage}[t]{0.2\linewidth}
			\centering
			\raisebox{0.0\height}{\includegraphics[width=1\linewidth]{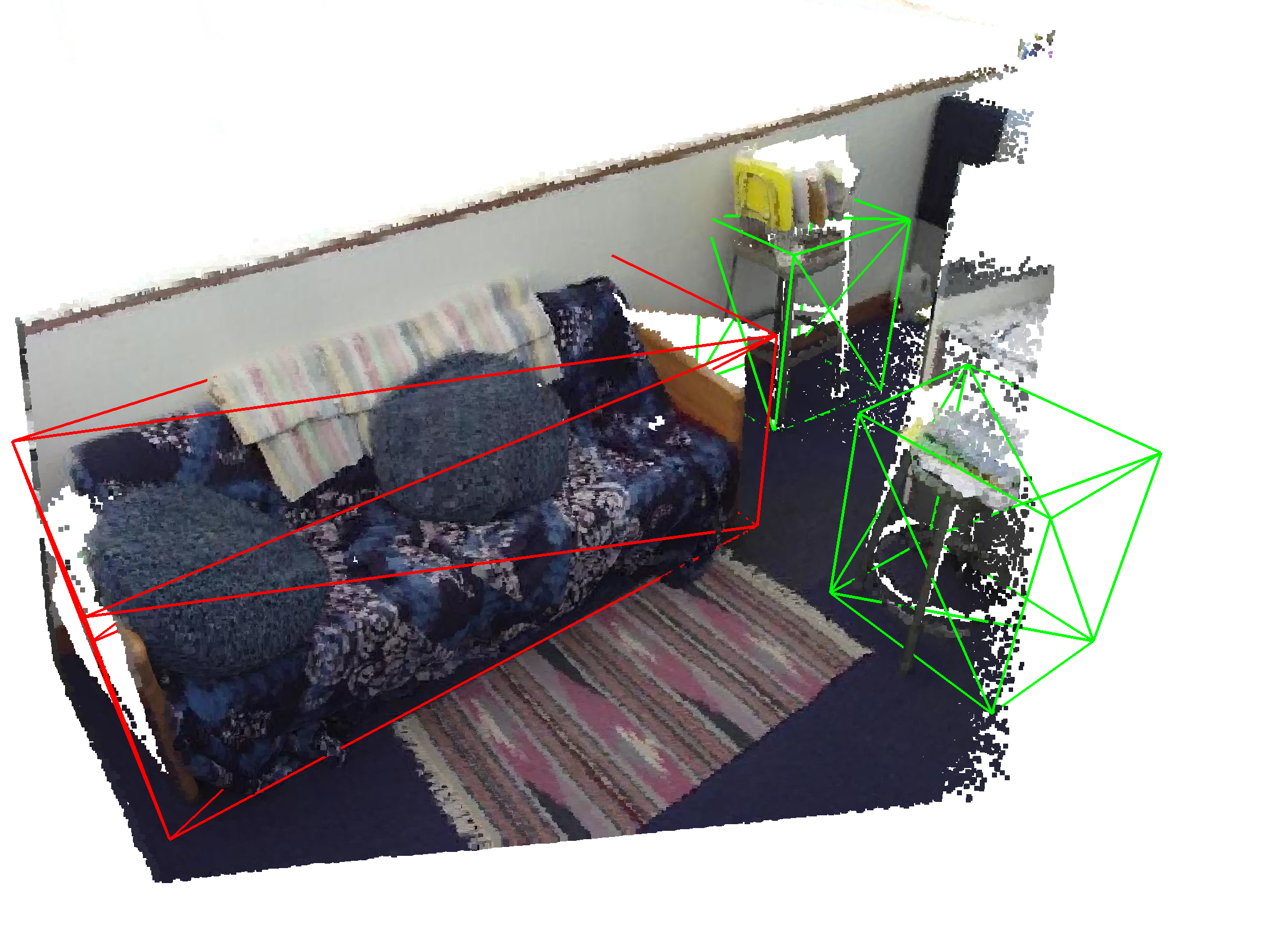}}
		\end{minipage}
	}
	\subfigure[Scene 3]{
		\begin{minipage}[t]{0.2\linewidth}
			\centering
			\raisebox{0.0\height}{\includegraphics[width=1\linewidth]{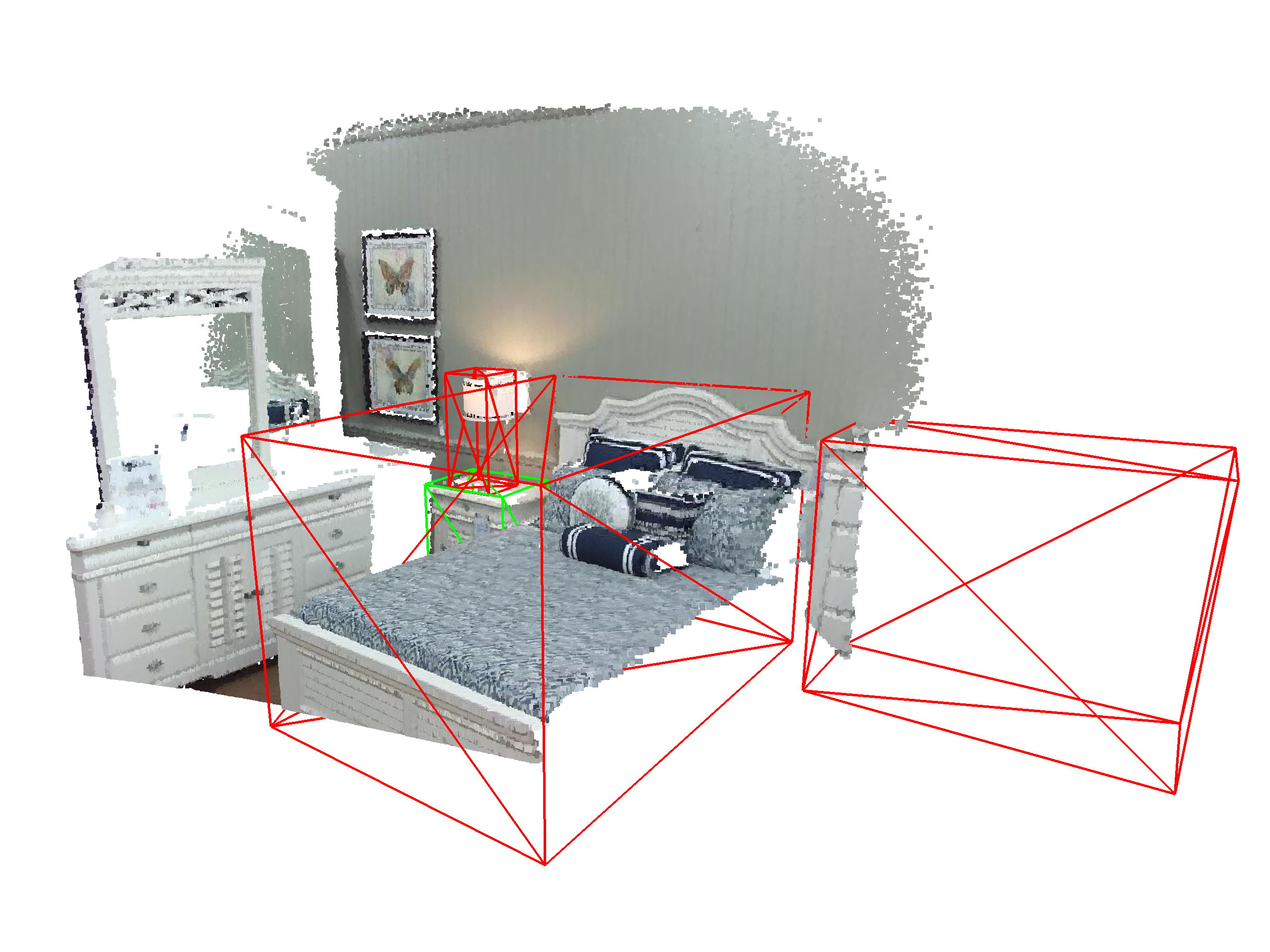}}
		\end{minipage}
	}
	\subfigure[Scene 4]{
		\begin{minipage}[t]{0.2\linewidth}
			\centering
			\raisebox{0.0\height}{\includegraphics[width=1\linewidth]{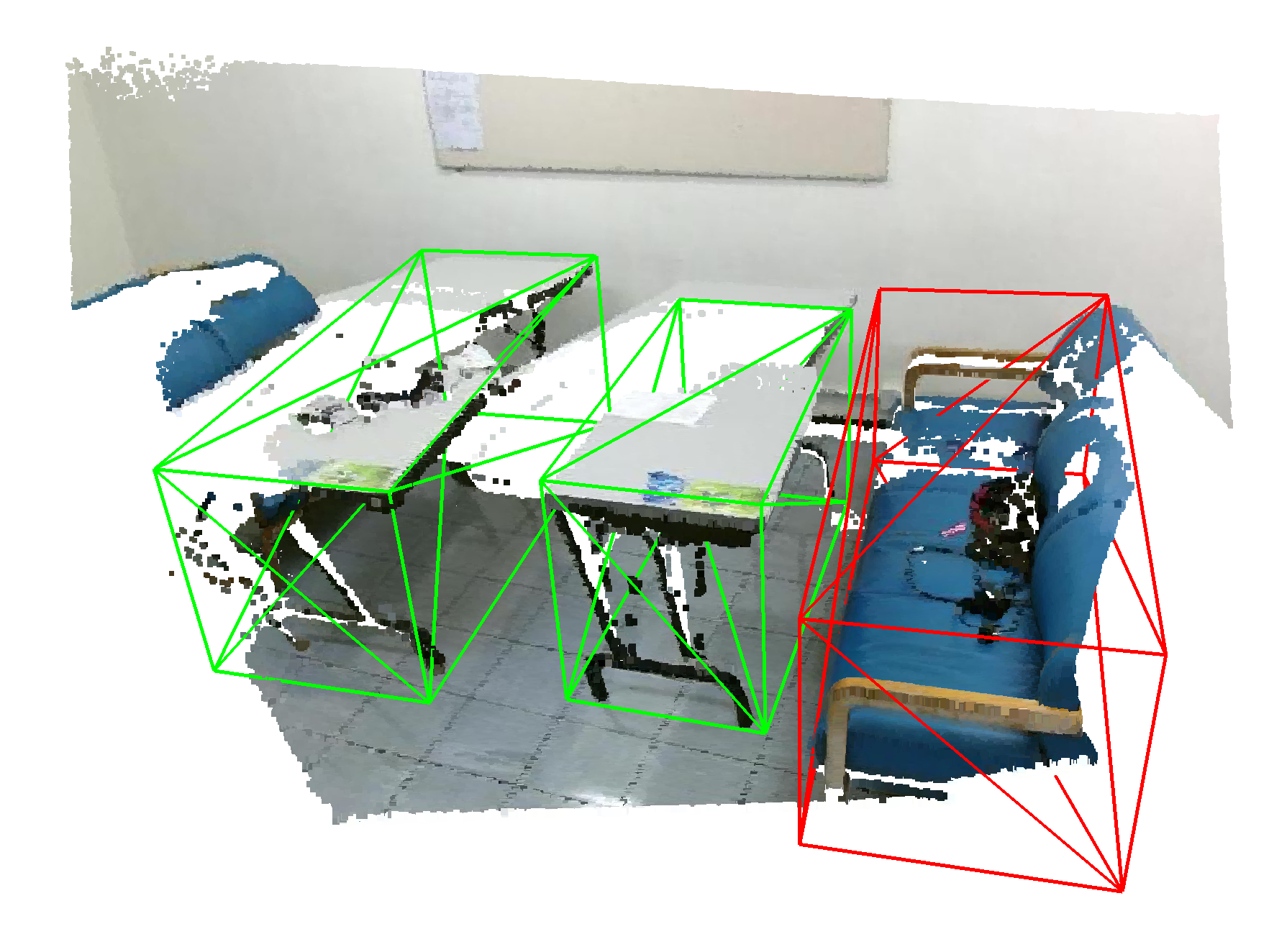}}
		\end{minipage}
	}
	
	\caption{\textbf{Visualization of detection results}. Four columns represent four different scenes, the comparison conducts among \textbf{Baseline}, \textbf{OV-3DETIC} and \textbf{Ground Truth}. All the bounding boxes in Baseline and OV-3DETIC are predictions, while the red bounding boxes in Ground Truth represent unseen class samples, and the green represents seen classes.}
	\label{fig:case_visualization}
\end{figure*}

Figure. \ref{fig:case_visualization} presents four cases of detection results. We can observe that the baseline is able to generate relatively accurate locations compared with ground truth, but the size, center, and direction are incorrect, especially the unseen class bounding boxes (chair in Scene 1, sofa in scene 2, bed in scene 3 and sofa in scene 4). Compare baseline with OV-3DETIC, OV-3DETIC performs much better on the unseen classes.

\section{Conclusion}
In this paper, we study a new problem of open-vocabulary 3D detection. The proposed method OV-3DETIC introduces ImageNet1K to help openv-vocabulary point-cloud detection. 
OV-3DETIC consists of two components: 1) we take advantage of two modalities --- the image modality for classification and the point-cloud modality for localization, to generate pseudo labels for unseen classes, and 2) debiased cross-modal contrastive learning to transfer the knowledge from images to point-clouds. 
Extensive experiments show that we improve a wide range of baselines by a large margin, demonstrating the effectiveness of the proposed method. We also provide explanations of why it works via ablation studies and analyzing the representations. As far as we know, we are the first to study open-vocabulary 3D detection, we do hope our work could inspire the research community to further explore this field.


{
\normalem
\bibliographystyle{unsrt}
\bibliography{ref}
}

\end{document}